\documentclass[lettersize,journal]{IEEEtran}
\usepackage{amsmath,amsfonts}
\usepackage{algorithmic}
\usepackage{algorithm}
\usepackage{array}
\usepackage[caption=false,font=normalsize,labelfont=sf,textfont=sf]{subfig}
\usepackage{textcomp}
\usepackage{stfloats}
\usepackage{url}
\usepackage{verbatim}
\usepackage{graphicx}
\usepackage{cite}
\usepackage{bm}
\usepackage{hyperref}

\usepackage{threeparttable}
\usepackage{booktabs}       % professional-quality tables
\usepackage{multirow} 
\usepackage{amssymb}

\usepackage{pifont}
\usepackage{color} % (if used in text)

\usepackage{xcolor}
\usepackage{colortbl,booktabs}

\newcommand{\cmark}{\ding{51}} 
\newcommand{\xmark}{\ding{55}}

\hypersetup{
	colorlinks=true,
	citecolor=cyan ,
    linkcolor=red,
}

\begin{document}

\title{EDA-DM: Enhanced Distribution Alignment for Post-Training Quantization of Diffusion Models}

\author{Xuewen Liu, Zhikai Li, Junrui Xiao, Mengjuan Chen, Jianquan Li, and Qingyi Gu,~\IEEEmembership{Senior Member,~IEEE}
        % <-this % stops a space ~\IEEEmembership{Graduate Student Member,~IEEE,}
\thanks{This work is supported in part by the National Natural Science Foundation of China under Grant 62276255; in part by the National Key Research and Development Program of China under Grant 2022ZD0119402. (Corresponding author: Zhikai Li, Qingyi Gu.)}% <-this % stops a space

\thanks{Xuewen Liu, Zhikai Li, and Junrui Xiao are with the Institute of Automation, Chinese Academy of Sciences, Beijing 100190, China, and also with the School of Artificial Intelligence, University of Chinese Academy of Sciences, Beijing 100049, China (e-mail: liuxuewen2023@ia.ac.cn; lizhikai2020@ia.ac.cn; xiaojunrui2020@ia.ac.cn).}

\thanks{Mengjuan Chen, Jianquan Li, and Qingyi Gu are with the Institute of Automation, Chinese Academy of Sciences, Beijing 100190, China (e-mail: mengjuan.chen@ia.ac.cn; jianquan.li@ia.ac.cn; qingyi.gu@ia.ac.cn).}
}

\markboth{Journal of \LaTeX\ Class Files,~Vol.~14, No.~8, August~2021}%
{Shell \MakeLowercase{\textit{et al.}}: A Sample Article Using IEEEtran.cls for IEEE Journals}


\maketitle

\begin{abstract}
    Diffusion models have achieved great success in image generation tasks. 
    However, the lengthy denoising process and complex neural networks hinder their low-latency applications in real-world scenarios. 
    Quantization can effectively reduce model complexity, and post-training quantization (PTQ), which does not require fine-tuning, is highly promising for compressing and accelerating diffusion models.
    Unfortunately, we find that due to the highly dynamic activations, existing PTQ methods suffer from distribution mismatch issues at both calibration sample level and reconstruction output level, which makes the performance far from satisfactory.
    In this paper, we propose EDA-DM, a standardized PTQ method that efficiently addresses the above issues.
    Specifically, at the calibration sample level, we extract information from the density and diversity of latent space feature maps, which guides the selection of calibration samples to align with the overall sample distribution; and at the reconstruction output level, we theoretically analyze the reasons for previous reconstruction failures and, based on this insight, optimize block reconstruction using the Hessian loss of layers, aligning the outputs of quantized model and full-precision model at different network granularity.
    Extensive experiments demonstrate that EDA-DM significantly outperforms the existing PTQ methods across various models and datasets. Our method achieves a 1.83$\times$ speedup and 4$\times$ compression for the popular Stable-Diffusion on MS-COCO, with only a 0.05 loss in CLIP score. Code is available at \href{http://github.com/BienLuky/EDA-DM}{http://github.com/BienLuky/EDA-DM}
\end{abstract}


\section{Introduction}

\IEEEPARstart{D}{iffusion models}~\cite{sohl2015deep,ho2020denoising,niu2020permutation} have gradually gained prominence in image generation tasks. Both considering the quality and diversity, they can compare or even outperform the SoTA GAN models~\cite{ozbey2023unsupervised}.
Furthermore, the flexible extensions of diffusion models achieve great performance in many downstream tasks, such as super-resolution~\cite{10685062}, image inpainting~\cite{lugmayr2022repaint}, motion prediction~\cite{10566611}, style transfer~\cite{10494219}, text-to-image~\cite{saharia2022photorealistic,luo2023latent}, and text-to-video~\cite{yang2024cogvideox,khachatryan2023text2video}.

Nevertheless, since diffusion models iteratively denoise the input using the same network within a single inference, the lengthy denoising process and complex neural networks hinder their low-latency applications in real-world scenarios.
To accelerate diffusion models, previous works~\cite{nichol2021improved,song2020denoising,10738304,lu2022dpm,liu2022pseudo} have focused on finding shorter and more efficient generation trajectories, thus reducing the number of steps in the denoising process.
Unfortunately, the complex network they ignored is also an important factor that consumes high memory and slows down the model at each denoising step.
For instance, even in the A6000 GPU, Stable-Diffusion~\cite{rombach2022high} still takes over a second to perform one denoising step with 16GB GPU memory.

Quantization techniques not only accelerate network, but also reduce the model memory footprint, which are extremely beneficial for generalizing diffusion models in low-latency applications.
Recently, model quantization includes two main approaches: quantization-aware training (QAT)~\cite{gong2019differentiable,nagel2022overcoming} and post-training quantization (PTQ)~\cite{li2023repq,nagel2020up}. 
While QAT can maintain performance by fine-tuning the whole models, it requires a significant amount of training data and expensive resources.
For instance, TDQ~\cite{so2024temporal} retrains DDIM~\cite{song2020denoising} on CIFAR-10~\cite{krizhevsky2009learning} using a 50K original dataset with 200K iterations. EfficientDM~\cite{he2023efficientdm} utilizes an additional LoRA~\cite{hu2021lora} module to fine-tune DDIM on CIFAR-10 with 12.8K iterations and 819.2K samples.
On the other hand, PTQ exhibits efficiency in terms of both data and time usage, which is more desired for compressing diffusion models.

\begin{figure}[t]
    \centering
    % \vspace{0.5cm}
    \includegraphics[width=1.0\linewidth]{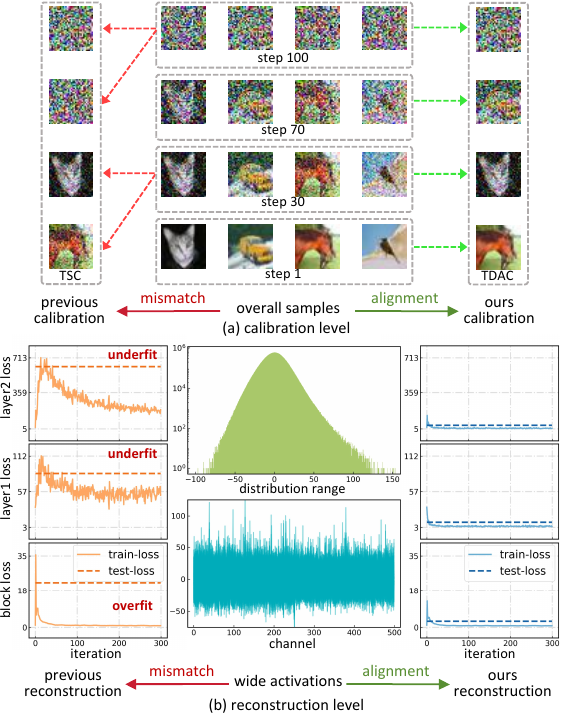}
    \vspace{-0.5cm}
    \caption{Visualization of the distribution mismatch at two levels for diffusion model quantization. (a) The temporal activations result in mismatch between the previous calibration and the overall samples. (b) The wide range of activations result in overfitting and underfitting in previous reconstruction.}
    \label{fig:motivation}
    \vspace{-0.5cm}
\end{figure}

PTQ generally follows a simple pipeline: obtaining calibration samples and then reconstructing the output.
However, as shown in Fig.~\ref{fig:motivation}, previous PTQ methods fail in diffusion models because the highly dynamic activations lead to a distribution mismatch at two levels:
\textbf{1) At calibration sample level}, since the diffusion models have an iterative denoising process, the input samples changed with time steps result in temporal activations, making it difficult to align calibration samples with the overall sample distribution.
Previous methods~\cite{shang2023post,li2023q,wang2023towards} select calibration samples based on experiment and observation. However, these methods are suboptimal or introduce computational overhead.
\textbf{2) At reconstruction output level}, the activations in diffusion models have a wide range, which increases the difficulty of quantization.
Using the previous reconstruction methods results in the outputs mismatch between the quantized model and the full-precision model.
Specifically, block-wise reconstruction~\cite{li2021brecq} over-enhances the dependence within the block layers resulting in overfitting, while layer-wise reconstruction~\cite{hubara2020improving} ignores the connections across layers resulting in underfitting.

To address the above issues, we propose a novel PTQ method for diffusion models, EDA-DM, which improves the performance of quantization at two levels.
At the calibration sample level, we extract information from the feature maps in the latent space for guiding the selection of calibration samples.
Based on the density and variety of feature maps, we propose \textit{\textbf{Temporal Distribution Alignment Calibration (TDAC)}} that effectively aligns the distribution of the calibration samples with that of the overall samples. 
At the reconstruction output level, we propose \textit{\textbf{Fine-grained Block Reconstruction (FBR)}}, which optimizes the loss of a block by incorporating the losses of layers within the block. This approach mitigates over-dependence within the block and enhances the connections between layers, aligning the outputs of quantized models and full-precision models at different network granularity.
To the best of our knowledge, existing PTQ methods for diffusion models ignore the effect of reconstruction, while this is the first work to analyze and improve the reconstruction method based on the properties of diffusion models.
Besides, our method does not introduce additional overhead or rely on a large number of quantization parameters, ensuring the deployment efficiency.
We also deploy the quantized diffusion models on different hardware platforms (GPU, CPU, ARM) to visualize the effect of quantization techniques on the compression and acceleration for diffusion models.
Overall, our contributions are summarized as follows:
\begin{itemize}
    \item Through thorough analysis, we identify two levels of mismatch in diffusion models, including the calibration sample level and the reconstruction output level, which result in the low performance of PTQ.
    \item Based on the above insight, we propose EDA-DM, an efficient PTQ method for compressing and accelerating diffusion models. Specifically, we propose TDAC to address the calibration sample level mismatch, and propose FBR to eliminate the reconstruction output level mismatch.
    \item Extensive results show that EDA-DM significantly outperforms the existing PTQ methods, especially in low-bit cases. Additionally, EDA-DM demonstrates robustness across various factors, such as model scale, resolution, guidance conditions, sampler, and hyperparameters.
\end{itemize}
\section{Related work}

\subsection{Efficient Diffusion Models}
\begin{figure}[!t]
    \centering
    \includegraphics[width=1.0\linewidth]{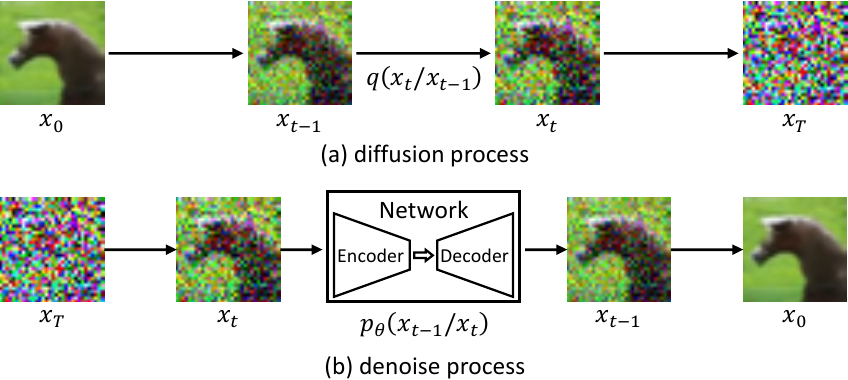}
    \caption{Brief illustration of Diffusion Model. In the training, the diffusion process (a) gradually adds noise to the real data $x_0$. In the inference, the denoising process (b) iteratively uses the network to denoise noise from Gaussian data $x_T$.}
    \label{fig:dm}
    \vspace{-0.3cm}
\end{figure}

Diffusion models have been proposed in 2015~\cite{sohl2015deep} and applied to image generation in 2020~\cite{ho2020denoising}, which consists of two processes. As shown in Fig.~\ref{fig:dm}, the forward diffusion process gradually adds noise to real data $x_0$, generating isotropic Gaussian data $x_T$. The denoising process removes the noise from the input $x_T$ step by step, generating the target image, where the noise is typically estimated by the UNet~\cite{ronneberger2015u} network or transformer~\cite{Peebles2022DiT} network.
While diffusion models~\cite{ho2020denoising,niu2020permutation} have generated high-quality images, the lengthy iterative denoising process and complex neural networks hinder their applications in real-world scenarios. 
Recently, efficient diffusion models have become a key focus of research in the community.
To shorten the lengthy denoising process, DDPM~\cite{nichol2021improved} adjusts the variance schedule; DDIM~\cite{song2020denoising} and BPA~\cite{10738304} generalizes diffusion process to a non-Markovian process with fewer denoising steps; PLMS~\cite{liu2022pseudo} and DPM~\cite{lu2022dpm} derive high-order solvers to approximate diffusion generation;
Deepcache~\cite{ma2024deepcache} and $\Delta$-DiT~\cite{chen2024delta} use cache mechanism to reduce the inference path at each step.
Distillation-based approaches optimize from two perspectives to accelerate diffusion models. Some methods~\cite{salimans2022progressive,luo2023latent} distill the generative capability of multiple denoising steps into fewer steps, while others~\cite{Lu2022KnowledgeDO,kim2023bk} design more lightweight noise estimation networks. 
On the other hand, compression-based methods improve inference speed by simplifying the complex neural networks of diffusion models.
For instance, LAPTOP-Diff~\cite{zhang2024laptop} and Diff-Pruning~\cite{fang2023structural} applies structured pruning to the pre-trained network, while Q-Diffusion~\cite{li2023q} and DilateQuant~\cite{liu2024dilatequant} quantize the network to lower bit precision.

\subsection{Quantization of Diffusion Models}
Several methods have been proposed for quantization of diffusion models.
Based on whether the model weights require retraining, these methods are generally fall into two categories: 
(1) Quantization-Aware Training (QAT).
TDQ~\cite{so2024temporal} and DilateQuant~\cite{liu2024dilatequant} retrains both the quantization parameters and weights. 
EfficientDM~\cite{he2023efficientdm} fine-tunes all of the model’s weights with an additional LoRA~\cite{hu2021lora} module, while QuEST~\cite{wang2024quest} selectively trains some sensitive layers. 
Although these methods can maintain the performance of quantized models, they require a significant amount of training data and expensive resources.
(2) Post-Training Quantization (PTQ). Compared to QAT, PTQ exhibits efficiency in terms of data and resource usage, as it does not require fine-tuning of model weights.
PTQ4DM~\cite{shang2023post} and Q-Diffusion~\cite{li2023q} design specific calibration samples based on observation and empirical evidence.
APQ-DM~\cite{wang2023towards} obtains calibration samples based on search algorithm, which introduces computational overhead. 
Additionally, it employs 8$\times$ quantization parameters to mitigate the dynamic nature of activation. 
TFMQ-DM~\cite{huang2024tfmq} further extends this approach by assigning different quantization parameters to each denoising step.
PTQD~\cite{he2023ptqd} uses statistical methods to estimate the quantization error, while TAC-Diffusion~\cite{yao2024timestep} propose a timestep-aware correction to dynamically corrects the quantization error.
TCAQ-DM~\cite{huang2024tcaq} employs reparameterization to reduce the difficulty of quantization.
Although these methods have achieved remarkable success, they also come with certain limitations. Some approaches~\cite{wang2023towards,yao2024timestep,huang2024tcaq} introduce additional computational overhead during inference. Others~\cite{so2024temporal,liu2024dilatequant,he2023efficientdm,wang2024quest,wang2023towards,huang2024tfmq,huang2024tcaq} set a large number of quantization parameters. 
These additional operations reduce the efficiency of quantized model deployment.
In contrast, we propose a standardized PTQ method that further enhances performance while maintaining hardware-friendly deployment.

\section{Methodology}
We start by detailing diffusion models and quantization techniques in Sec.~\ref{sec:3.0}, then explore the challenges of PTQ for diffusion models in Sec.~\ref{sec:3.1}, and finally propose our efficient methods to address these challenges in Sec.~\ref{sec:3.2} and Sec.~\ref{sec:3.3}.

\subsection{Preliminary}\label{sec:3.0}
\subsubsection{Diffusion Model} As shown in Fig.~\ref{fig:dm}, in the training, the forward diffusion process gradually adds Gaussian noise to real data $x_0\sim q\left( x_0\right)$ for $T$ times, which is a Markov process:
\begin{align}
    q\left(x_t \mid x_{t-1}\right)=\mathcal{N} (x_t ; \sqrt{1-\beta_t} x_{t-1}, \beta_t \bm{I}) 
\end{align}
where $\beta_t$ is the hyperparameter. When $T$ is sufficiently large $T\sim \infty $, $x_T$ approximates an isotropic Gaussian distribution $x_T\sim \mathcal{N}\left( 0, \bm{I}\right)$.

In the inference, the denoising process removes the noise from the input $x_T$ to generate high-quality images. Since $q\left(x_{t-1} \mid x_t\right)$ relies on $q\left( x_0\right)$, which is unavailable, diffusion model approach it by learning a Gaussian distribution:
\begin{align}
    p_\theta\left(x_{t-1} \mid x_t\right)=\mathcal{N}\left(x_{t-1} ; \mu_\theta\left(x_t, t\right), \Sigma_\theta\left(x_t, t\right)\right) 
\end{align}
where the variance $\Sigma_\theta\left(x_t, t\right)$ can be fixed as a constant schedule $\sigma_t$ to make the training stable. And with the reparameterization trick~\cite{ho2020denoising}, the mean $\mu_\theta\left(x_t, t\right)$ can be formulated as:
\begin{align}
    \mu_\theta\left(x_t, t\right)=\frac{1}{\sqrt{\alpha_t}}( x_t-\frac{\beta_t}{\sqrt{1-\bar{\alpha}_t}} \epsilon_\theta\left(x_t, t\right) )
\end{align}
where $\alpha_t=1-\beta_t$, $\bar{\alpha}_t= {\textstyle \prod_{k=1}^{t}} \alpha_k$. Finally, the denoising process generates $x_{t-1}$ by predicting $\epsilon_\theta\left(x_t, t\right)$ through the noise estimation network:
\begin{align}
    x_{t-1}=\frac{1}{\sqrt{\alpha_t}}( x_t-\frac{\beta_t}{\sqrt{1-\bar{\alpha}_t}} \epsilon_\theta\left(x_t, t\right) )+\sigma_t z
\end{align}
where $z=\mathcal{N}\left( 0, \bm{I}\right)$. As can be seen, diffusion models iteratively denoise the input using the same network within a single inference.
The variation of network input samples across time steps results in a highly dynamic distribution of activations.

\subsubsection{Model Quantization}
Quantization transforms the floating-point value $x$ of weights and activations to quantized value $\hat{x}$ using the quantization parameters: scale factor $s$ and zero point $z$. The uniform quantizer used in our work can be formulated as: 
\begin{align}\label{eq:quant}
    \bar{x}=clip \left( \left\lfloor\frac{\bm{x}}{s}\right\rceil+z, 0, 2^b-1 \right), \;\hat{x}=s \cdot ( \bar{x}-z ) 
\end{align}
where $\left\lfloor{\cdot}\right\rceil$ represents rounding opration, the bit-width $b$ determines the range of clipping function $clip(\cdot )$, and the $\bar{x}$ is integer value for hardware efficiency. 

To set the appropriate quantization parameters, 
PTQ typically follows two processes: obtaining the calibration samples and reconstructing the model output. 
The calibration samples characterize the overall samples to calibrate the quantization parameters. 
On the other hand, Reconstruction process utilizes distillation techniques to align the outputs of quantized models and full-precision models.
The most widely used block-wise reconstruction with loss as:
\begin{align}
    L_b = {\arg \min } {\left \| \hat{\bm{y}}(x) - \bm{y}(x) \right \| }_F^2
\end{align}
has demonstrated success in classification and detection networks~\cite{xiao2023patch,li2023vit,li2023psaq}, where $\hat{\bm{y}}(x)$ and $\bm{y}(x)$ represent the outputs of the quantized model and full-precision model at one block, respectively, ${\left \| \cdot \right \| }_F^2$ denotes Frobenius Norm.
However, due to the highly dynamic activations caused by the unique temporal denoising process, previous PTQ methods suffer from severe performance degradation for diffusion models.
Existing PTQ methods introduce additional overhead or a large number of quantization parameters to recover accuracy. 
This results in the inefficient deployment of the quantized models.

\subsection{Challenges of PTQ for Diffusion Models}\label{sec:3.1}
We revisit the challenges of PTQ for diffusion models.
Through experiments, we find that the highly dynamic distribution of activations results in two levels of mismatch, making the quantization worse.
Specifically, the temporal nature of activations results in calibration sample level mismatch and the wide range of activations results in reconstruction output level mismatch.

\begin{figure*}[t]
    \centering
    \includegraphics[width=0.95\textwidth]{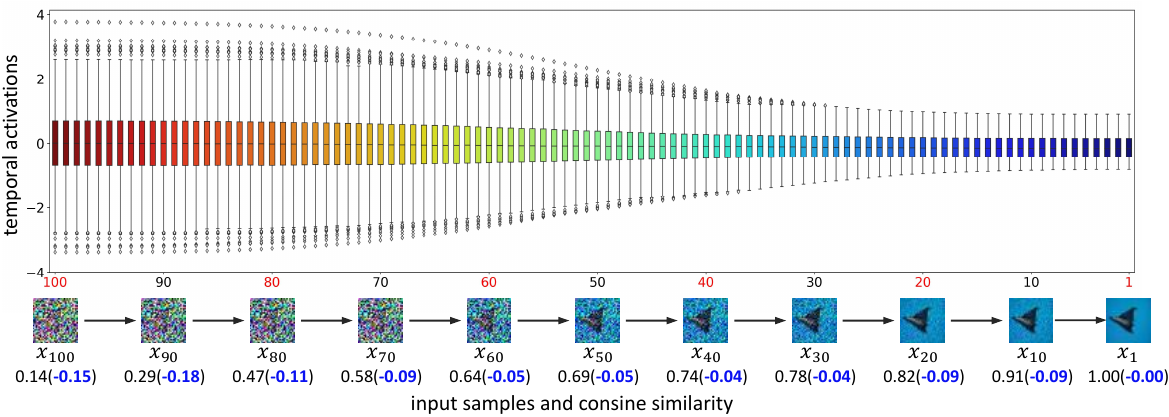}
    \caption{An overview of the challenge 1. The cosine similarity is obtained by calculating the cosine distance of feature maps between $x_t$ and $x_1$. As observed, the cosine similarity decreases at varying rates every 10 time steps, indicating that sample variations are not uniform. Data comes form DDIM on CIFAR-10.}
    \label{fig:challenge1}
    \vspace{-0.3cm}
\end{figure*}
\subsubsection{Challenge 1: Calibration Sample Level Mismatch}
The calibration samples are expected to characterize the overall sample distribution, which helps to reduce quantization errors.
For diffusion models with the temporal denoising process, at any time step $t$, the output sample $x_t$ is the input of denoising network at time step $t-1$ (as shown in Fig.~\ref{fig:dm}). This causes the inputs of the network to change with each time step, resulting in activations that exhibit a temporal nature, as shown in Fig.~\ref{fig:challenge1}. The calibration samples for diffusion models require to align with the overall samples to reduce quantization errors at all time steps.

Existing research~\cite{shang2023post,li2023q,wang2023towards} on calibration for diffusion models remains focused at the sample level, where calibration is designed based on variations among input samples.
PTQ4DM~\cite{shang2023post}, based on empirical observations, constructs the calibration by sampling from different time steps according to a skew-normal distribution ratio. 
Q-Diffusion~\cite{li2023q} obtains the calibration using a uniform time-step sampling strategy (TSC), as shown in Fig.~\ref{fig:motivation}.
ADP-DM~\cite{wang2023towards} employs an optimization-based approach to obtain the calibration from a single time step. However, it introduces additional computational overhead and an 8$\times$ increase in quantization parameters.
Considering time and resource efficiency, TSC has been adopted by other quantization methods~\cite{wang2024quest,huang2024tfmq}.
This sampling strategy is based on a strong assumption: \textit{sample variations at different time steps remain consistent}.
To validate this assumption, we characterize sample variations by calculating the cosine similarity of the network feature maps, which have been demonstrated to effectively represent sample distribution~\cite{chen2021image, park2018accelerating}.
Unfortunately, our findings reveal that: \textit{sample variations are not uniform at different time steps}. More specifically, sample variation is pronounced in the initial and final time steps, while it is significantly reduced during the intermediate time steps.
Therefore, at the calibration sample level, previous sampling strategies are not optimal, and an efficient and rational sampling strategy is required.

\begin{figure*}[!t]
    \centering
    \includegraphics[width=0.95\textwidth]{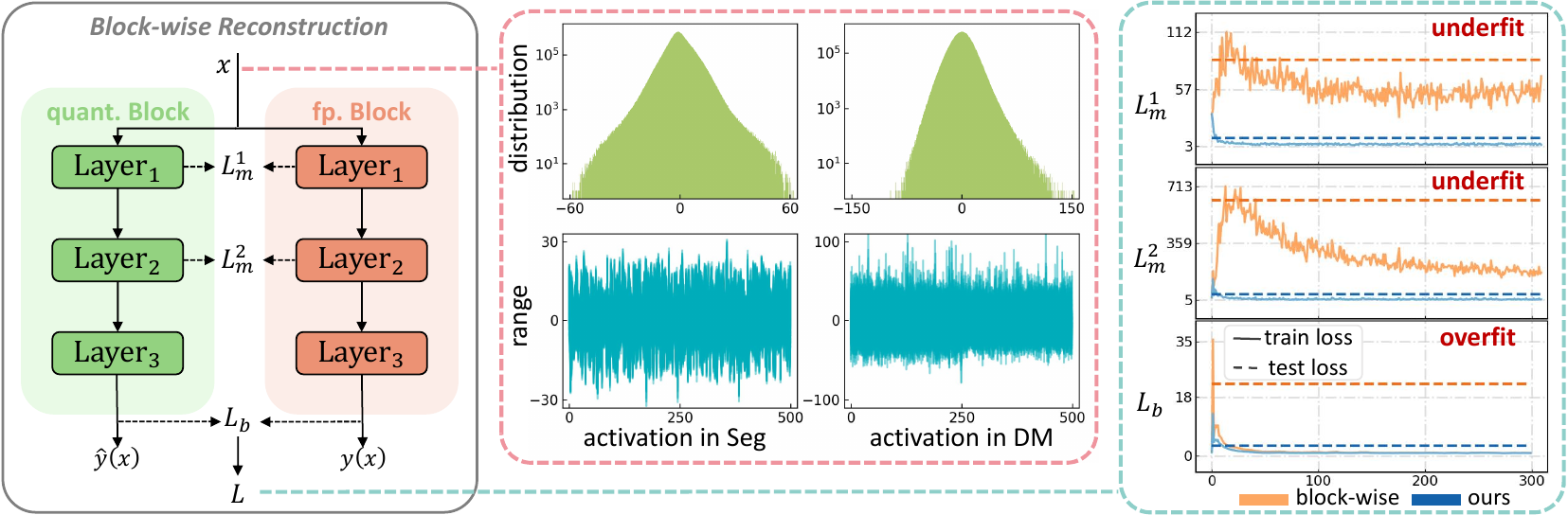}
    \caption{An overview of the challenge 2. The data and losses are obtained from the last Residual Bottleneck Block of the middle stage of the UNet network.}
    \label{fig:challenge2}
    \vspace{-0.3cm}
\end{figure*}
\subsubsection{Challenge 2: Reconstruction Output Level Mismatch}
Reconstruction is a crucial method for enhancing quantization performance, especially in low-bit cases.
For single time step models, previous works have already demonstrated that block-wise reconstruction~\cite{li2021brecq} can balance the cross-layer dependency and generalization error, resulting in superior quantization performance.
However, when applying this approach to diffusion models, the performance is far from satisfactory.

To explore the reasons thoroughly, we quantize DDIM to 4-bits with block-wise reconstruction and examine the reconstruction performance of blocks and layers within the blocks.
As shown in Fig.~\ref{fig:challenge2}, the activations in diffusion models have a wide range, making them hard to quantize.
For example, in the same Residual Bottleneck Block of UNet networks, the range of activations in diffusion models is almost 3$\times$ larger than that in segmentation model~\cite{10173725}.
To align the quantized block with the full-precision block, block-wise reconstruction struggles to decrease the block loss $L_b$ at the expense of increasing the losses of the front layers ($L^{(1)}_{m}, L^{(2)}_{m}$).
As a result, the reconstructed block is overfitted, and the front layers are underfitted. Namely, the output of reconstruction is mismatched.
To preserve time-step guidance information, TFMQ-DM~\cite{huang2024tfmq} separates the {\tt{embedding layer}} from the block and reconstructs it independently. However, this approach only mitigates quantization errors in the {\tt{embedding layer}} and fails to address overfitting in block and underfitting in other layers.

\subsection{Temporal Distribution Alignment Calibration}\label{sec:3.2}
To address the calibration sample level mismatch, we attempt to extract information from the temporal network to guide the selection of calibration samples. Feature map is a mapping of network inputs into the latent space, encompassing the feature and distribution information of input samples~\cite{chen2021image, park2018accelerating}.
In this work, we utilize the output of the middle stage of the network as a feature map because it contains high-dimensional information of the input samples~\cite{ma2024deepcache}.
Since the diffusion model runs the network $T$ times in one inference, we obtain feature maps from each time step to form $F={\left \{ F_t \right \} }_{t=1}^T$.
Based on the set $F$, we propose \emph{\textbf{Density score}} $D={\left \{ D_t \right \} }_{t=1}^T$, which effectively quantifies the ability of each time-step input samples to represent the overall samples.
Furthermore, given that hard samples significantly influence the quantization~\cite{li2023hard}, we introduce \emph{\textbf{Variety score}} $V={\left \{ V_t \right \} }_{t=1}^T$, which quantifies the diversity of each time-step input samples.
The effectiveness of the two scores is demonstrated in Sec.~\ref{sec:score}.
For $t_{th}$ time step, the mathematical formulas for $D_t$ and $V_t$ are as follows:
\begin{align}
    D_t &= \Big | \left \{ {F_i \,|\, mse\left ( {F_t,F_i} \right ) < \varepsilon}, F_i \in F  \right \} \Big | \\
    V_t &= \sum_{i=1}^T \left ( {1-dist \left ( {F_t,F_i} \right )} \right )
\end{align}
where the function $mse\left ( {\cdot} \right )$ calculates the MSE distances, $dist\left ( {\cdot} \right )$ calculates the cosine similarity.
% , and we also conduct the ablation study to demonstrate the effectiveness of the metric in Appendix~\ref{app:metric}.
The $\varepsilon$ represents the distance threshold, which is set as a fixed constant for all tasks, and the function $\big | {\cdot} \big |$ counts the number of set elements.
Namely, $D_t$ represents the density of the feature maps $F$ with respect to $F_t$, while $V_t$ denotes the dissimilarity between $F$ and $F_t$.
We use the \emph{Min-Max Scaling} to eliminate the effect of the magnitudes, obtaining the effective scores $\hat{D_t}$ and $\hat{V_t}$:
\begin{align}
    \hat{D_t} &= \frac{D_t-min(D)}{max(D)-min(D)} \label{eq:D} \\
    \hat{V_t} &= \frac{V_t-min(V)}{max(V)-min(V)} \label{eq:V}
\end{align}

\begin{figure}[!t]
    \centering
    \includegraphics[width=0.90\linewidth]{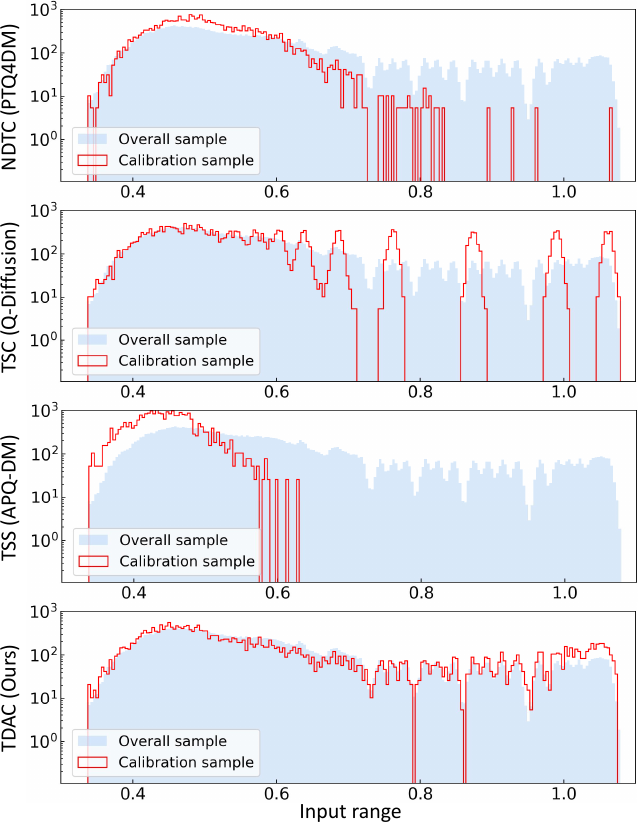}
    \vspace{-0.1cm}
    \caption{Visualization of the different sampling strategies. Here, the x-axis represents the distance of the samples from the geometric center of the overall samples, and the y-axis represents the number of distributed samples.}
    \label{fig:distribution}
    \vspace{-0.3cm}
\end{figure}

The sum of the two scores $S_t$ determines the proportion of samples extracted from the $t_{th}$ time step to calibration samples. Finally, the \textit{\textbf{T}emporal \textbf{D}istribution \textbf{A}lignment \textbf{C}alibration} (\textbf{TDAC}) is as follows:
\begin{align}
    S_t &= \hat{D_t} + \lambda \ast \hat{V_t} \label{eq:S} \\
    X_t &= \frac{S_t}{\sum_{t=1}^T S_t} \ast N \label{eq:X}
\end{align}
where hyperparameter $\lambda$ balances these two scores, and $N$ represents the number of calibration samples. $X_t$ denotes samples extracted from the $t_{th}$ time step, forming the calibration $X = {\left \{ X_t \right \} }_{t=1}^T$.
Consequently, compared to different sampling strategies, TDAC effectively addresses the mismatch in calibration sample levels, as shown in Fig.~\ref{fig:distribution}.
The overall pipeline of TDAC is shown in Fig.~\ref{fig:overview} (a).

\begin{figure*}[!th]
    \centering
    \includegraphics[width=0.96\textwidth]{image/overview.pdf}
    \vspace{-0.1cm}
    \caption{The overall pipeline of our method. TDAC addresses the calibration sample level mismatch by extracting information from the feature maps. FBR tackles the reconstruction output level mismatch by optimizing the reconstruction loss.}
    \label{fig:overview}
    \vspace{-0.3cm}
\end{figure*}

\begin{algorithm}[th]
    \caption{Overall quantization workflow of EDA-DM}
    \label{alg:algorithm}
    \textbf{Input}: Pre-trained full-precision model $W_r$ with $T$ steps.\\
    \textbf{Parameter}: The hyperparameters $\lambda$ and $\gamma$.\\
    \textbf{Output}: Quantized model $W_q$.
    
    \begin{algorithmic}[1] %[1] enables line numbers
        \STATE \textbf{TDAC:}
        \STATE Inference $W_r$ one time for obtaining the feature maps $F={\left \{ F_t \right \} }_{t=1}^T$ and input samples $x={\left \{ x_t \right \} }_{t=1}^T$.
        \FOR{$t = 1$ to $T$ time steps}
        \STATE Calculate the effective density score $\hat{D_t}$ and variety score $\hat{V_t}$ of $t_{th}$ input sample $x_t$ by Eq.~\ref{eq:D} and Eq.~\ref{eq:V}.
        \STATE Calculate the sum score $S_t$ of the $x_t$ by Eq.~\ref{eq:S}.
        % $S_t = D_t + \lambda \ast V_t$
        \ENDFOR
        \FOR{$x_t$ in input samples $x$}
        \STATE Calculate the proportion $X_t$ of calibration by Eq.~\ref{eq:X}.
        %$X_t = \frac{S_t}{\sum_{t=1}^T S_t} \ast N$
        \STATE Extract the $X_t$ samples from $x_t$, forming the TDAC.
        \ENDFOR
        \STATE Initialize the quantized model with TDAC.
        \STATE \textbf{FBR:}
        \FOR{$l = 1$ to the end block}
        \STATE Calculate the block loss $L_{b}$ and front layers losses $L_{m}$ for the $l_{th}$ block.
        \STATE Calculate the new loss $L$ by Eq.~\ref{eq:L}, and update quantizers by gradient descent algorithm.
        \ENDFOR
        \STATE \textbf{return} $W_q$
    \end{algorithmic}
\end{algorithm}

\subsection{Fine-grained Block Reconstruction}\label{sec:3.3}
We begin by analyzing the errors introduced by weight-activation quantization. For a linear layer with weights $\bm{W} \in \mathbb{R}^{m\times n}$, activations $\bm{x} \in \mathbb{R}^{n\times 1}$, and output $\bm{y} = \bm{W}\bm{x}$,$\bm{y} \in \mathbb{R}^{m\times 1}$, the quantization error can be expressed as:
\begin{align}
    E\left (\bm{W}, \bm{x} \right ) = \mathbb{E}_{\bm{x}\sim D_c } \left [  \mathcal{L}\left ( \hat{\bm{W}}, \hat{\bm{x}}\right ) - \mathcal{L}\left ( \bm{W}, \bm{x}\right )\right ] 
\end{align}
where $D_c$ denotes sample sets, $\hat{\bm{W}}$ and $\hat{\bm{x}}$ represent the quantized tensor.
According to the proof in appendix~\ref{sec:app}, quantized activation element $\hat{x}$ can be expressed as $\hat{x}=x\cdot \left ( 1+u\left ( x \right )  \right ) $, where $u$ is affected by bit-width and rounding error.
Consider matrix-vector multiplication, we have the quantized output $\hat{\bm{y}}=\hat{\bm{W}}\hat{\bm{x}}=\left (\hat{\bm{W}}\odot \left ( 1+\bm{V}\left ( \bm{x} \right )  \right )\right ) \bm{x}$, given by
\begin{align}
    \hat{\bm{W}}\hat{\bm{x}} &=
    \hat{\bm{W}}(\bm{x} \odot\left[\begin{array}{c}
    1+\bm{u}_1(\bm{x}) \\
    1+\bm{u}_2(\bm{x}) \\
    \ldots \\
    1+\bm{u}_n(\bm{x})
    \end{array}\right]) \\ 
    &= (\hat{\bm{W}} \odot\left[\begin{array}{cccc}
    1+\bm{u}_1(\bm{x}) & \ldots & 1+\bm{u}_n(\bm{x})\\
    1+\bm{u}_1(\bm{x}) & \ldots & 1+\bm{u}_n(\bm{x})\\
    \ldots & & \\
    1+\bm{u}_1(\bm{x}) & \ldots & 1+\bm{u}_n(\bm{x})
    \end{array}\right]) \bm{x}
\end{align}

As can be seen, by taking $\bm{V}_{i,j}\left ( \bm{x} \right )=\bm{u}_{j}\left ( \bm{x} \right )$, quantization error on the activation vector $\left(1+\bm{u}(\bm{x})\right)$ can be transplanted into perturbation on weight $\left(1+\bm{v}(\bm{x})\right)$. 
Thus, the error caused by weight-activation quantization can be briefly expressed as:
\begin{align}\label{eq:4.1}
    E\left (\bm{W}, \bm{x} \right ) = \mathbb{E}_{\bm{x}\sim D_c } \left [  \mathcal{L}\left ( \tilde{\bm{W}}, \bm{x}\right ) - \mathcal{L}\left ( \bm{W}, \bm{x}\right )\right ] 
\end{align}
where $\tilde{\bm{W}}=\hat{\bm{W}}\odot \left ( 1+\bm{V}\left ( \bm{x} \right )  \right )$.

Next, we perform a Taylor expansion on Eq.~\ref{eq:4.1}, approximating the quantization error as:
\begin{align}
    E\left (\bm{W}, \bm{x} \right ) \approx \Delta\bm{W}^T\bar{g}^{\left ( \mathbf{W} \right ) } + \frac{1}{2}\Delta\bm{W}^T \mathbf{H}^{\left ( \mathbf{W} \right ) }\Delta\bm{W}
\end{align}
by setting $\tilde{\bm{W}} = \bm{W}+\Delta \bm{W}$ and neglecting the impact of higher-order terms, 
where $\bar{g}^{\left ( \mathbf{W} \right ) }=\mathbb{E}\left[ \bigtriangledown_\mathbf{W} \mathcal{L} \right]$ and $\mathbf{H}^{\left ( \mathbf{W} \right )}=\mathbb{E}\left[ \bigtriangledown^2_\mathbf{W}\mathcal{L} \right]$ are the gradients and the Hessian matrix, $\Delta \bm{W}$ is the overall weight perturbation. 
Given the pretrained model is converged to a minimum, the gradients can be safely thought to be close to $\mathbf{0}$.
Therefore, the quantization error can be further expressed as:
\begin{align}\label{eq:4.2}
    E\left (\bm{W}, \bm{x} \right ) \approx \frac{1}{2}\Delta\bm{W}^T \mathbf{H}^{\left ( \mathbf{W} \right ) }\Delta\bm{W}
\end{align}

Optimizing Eq.~\ref{eq:4.2} by adjusting the quantization parameters can effectively reduce quantization error. This process is known as reconstruction.
However, optimizing with the large-scale full Hessian is memory-infeasible on many devices as the full Hessian requires terabytes of memory space.
Fortunately, existing theoretical studies~\cite{botev2017practical,li2021brecq} have demonstrated that optimizing the second-order error of the output can serve as an approximation for optimizing Eq.~\ref{eq:4.2}. Moreover, Adaround~\cite{nagel2020up} further shows that this can be approximated by minimizing the mean squared error (MSE) loss of the output:
\begin{align}
    \underset{\tilde{\bm{W}}}{\arg \min } \Delta \bm{W}^T {\mathbf{H}}^{(\mathbf{W})} \Delta \bm{W} &\approx \underset{\tilde{\bm{W}}}{\arg \min } \mathbb{E}\left[\Delta \bm{y}^T\mathbf{H}^{\left(\mathbf{y}\right)} \Delta \bm{y}\right] \nonumber \\
    &\approx \underset{\tilde{\bm{W}}}{\arg \min } {\left \| \hat{\bm{y}} - \bm{y} \right \| }_F^2
\end{align}

Then, we discuss reconstruction methods at different granularity.
Assuming the network consists of $n$ layers, layer-wise reconstruction~\cite{hubara2020improving} reconstructs the output layer by layer. For the $k_{th}$ layer, its reconstruction loss $L^{(k)}_m$ is expressed as: 
\begin{align}
    L^{(k)}_m=\underset{\tilde{\bm{W}}^{(k)}}{\arg \min } {\left \| \hat{\bm{y}}^{(k)} - \bm{y}^{(k)} \right \| }_F^2
\end{align}
here, the superscript $(k)$ denotes the tensors at the $k_{th}$ layer.
Layer-wise reconstruction completely ignores inter-layer dependency. Although it minimizes quantization errors on training set, it suffers from significant generalization errors on test set.
On the other hand, block-wise reconstruction~\cite{li2021brecq} reconstructs the output block by block.
Formally, if layer $k$ to layer $\ell $ (where $1 \le k < \ell \le n$) form a block, the weight vector is defined as $\tilde{\bm{\theta}} = vec[\tilde{\bm{W}}^{(k),T},\dots ,\tilde{\bm{W}}^{(\ell),T}]^T$ and the optimized Hessian is transformed by
\begin{align}
    \underset{\tilde{\bm{\theta}}}{\arg \min } \Delta \bm{\theta}^T {\mathbf{H}}^{(\mathbf{\theta})} \Delta \bm{\theta} 
    &\approx \underset{\tilde{\bm{\theta}}}{\arg \min } \mathbb{E}\left[\Delta \bm{y}^{(\ell),T}\mathbf{H}^{\left(\mathbf{y}^{(\ell)}\right)} \Delta \bm{y}^{(\ell)}\right] \nonumber \\
    &\approx \underset{\tilde{\bm{\theta}}}{\arg \min } {\left \| \hat{\bm{y}}^{(\ell)} - \bm{y}^{(\ell)} \right \| }_F^2
\end{align}
Its reconstruction loss $L_b$ denotes as: 
\begin{align}
    L_b=\underset{\tilde{\bm{\theta}}}{\arg \min } {\left \| \hat{\bm{y}}^{(\ell)} - \bm{y}^{(\ell)} \right \| }_F^2
\end{align}

Block-wise reconstruction ignores the inter-block dependency and considers the intra-block dependency.
Compared with layer-wise reconstruction, it balances quantization and generalization errors in networks with small-range activations, such as segmentation~\cite{nagel2020up} or classification networks~\cite{li2021brecq}.
However, diffusion models exhibit a wide range of activations, which leads to significant quantization errors in each layer within the block.
The severe quantization noise invalidates the assumption of complete intra-block dependency, rendering block-wise reconstruction ineffective in balancing quantization and generalization errors.
Specifically, the block is overfitted and the front layers are underfitted, as illustrated in Fig.~\ref{fig:challenge2}.
Therefore, when applying block-wise reconstruction to diffusion models, the performance is far from satisfactory.

Finally, since previous reconstruction methods fail in aligning the output at the reconstruction level, we propose \textit{\textbf{F}ine-grained \textbf{B}lock \textbf{R}econstruction} (\textbf{FBR}).
Our method reformulates the optimized Hessian as:
\begin{equation}
    \resizebox{1.0\hsize}{!}{$
    \begin{aligned}
        \underset{\tilde{\bm{\theta}}}{\arg \min } \mathbb{E}\left[\Delta \bm{y}^{(\ell),T}\mathbf{H}^{\left(\mathbf{y}^{(\ell)}\right)} \Delta \bm{y}^{(\ell)} + \gamma \cdot \sum_{i=k}^{\ell-1} \Delta \bm{y}^{(i),T}\mathbf{H}^{\left(\mathbf{y}^{(i)}\right)} \Delta \bm{y}^{(i)}\right] 
    \end{aligned}
    $}\nonumber
\end{equation}
\begin{equation}
    \resizebox{0.83\hsize}{!}{$
    \begin{aligned}
        \approx \underset{\tilde{\bm{\theta}}}{\arg \min } \left[{\left \| \hat{\bm{y}}^{(\ell)} - \bm{y}^{(\ell)} \right \| }_F^2 + \gamma \cdot \sum_{i=k}^{\ell-1} {\left \| \hat{\bm{y}}^{(i)} - \bm{y}^{(i)} \right \| }_F^2\right]
    \end{aligned}
    $}
\end{equation}
% \begin{equation}
%     \resizebox{1.0\hsize}{!}{$
%         \begin{aligned}
%             &\underset{\tilde{\bm{\theta}}}{\arg \min } \mathbb{E}\left[\Delta \bm{y}^{(\ell),T}\mathbf{H}^{\left(\mathbf{y}^{(\ell)}\right)} \Delta \bm{y}^{(\ell)} + \gamma \cdot \sum_{i=k}^{\ell-1} \Delta \bm{y}^{(i),T}\mathbf{H}^{\left(\mathbf{y}^{(i)}\right)} \Delta \bm{y}^{(i)}\right] \\
%             &\approx \underset{\tilde{\bm{\theta}}}{\arg \min } \left[{\left \| \hat{\bm{y}}^{(\ell)} - \bm{y}^{(\ell)} \right \| }_F^2 + \gamma \cdot \sum_{i=k}^{\ell-1} {\left \| \hat{\bm{y}}^{(i)} - \bm{y}^{(i)} \right \| }_F^2\right] 
%         \end{aligned}
%     $}
% \end{equation}
The new reconstruction Loss $L$ is expressed as:
\begin{align}\label{eq:L}
    L&=\underset{\tilde{\bm{\theta}}}{\arg \min } \left[{\left \| \hat{\bm{y}}^{(\ell)} - \bm{y}^{(\ell)} \right \| }_F^2 + \gamma \cdot \sum_{i=k}^{\ell-1} {\left \| \hat{\bm{y}}^{(i)} - \bm{y}^{(i)} \right \| }_F^2\right] \nonumber \\
    &=L_b + \gamma \cdot \sum_{i=k}^{\ell-1} L^{(i)}_m
\end{align}
where the hyperparameter $\gamma$ balances these two parts of the loss.
Obviously, FBR is able to reduce the quantization error across all layers within the block while preserving the generalization capability of the quantized block.
As depicted in Fig.~\ref{fig:challenge2}, ours FBR effectively eliminates overfitting of reconstructed blocks and underfitting of layers within blocks, aligning quantized models with full-precision models at the reconstruction output level. 
More importantly, it provides an efficient way to address the wide range activations in reconstruction process.
The overall EDA-DM workflow is presented in Algorithm~\ref{alg:algorithm}.

\section{Experiments}
\subsection{Implementation Details}
\subsubsection{Models and Datasets}
We evaluate EDA-DM on mainstream diffusion models (DDIM, LDM-4, LDM-8, Stable-Diffusion)~\cite{song2020denoising,rombach2022high} across six benchmark datasets (CIFAR-10, LSUN-Bedroom, LSUN-Church, ImageNet, MS-COCO, DrawBench)~\cite{krizhevsky2009learning,yu2015lsun,saharia2022photorealistic}.
All pre-trained models are obtained from their official sources.
For Stable-Diffusion, we quantize its v1.4 version.

We conduct all experiments on an RTX A6000 and deploy the quantized models on an RTX 3090 for real-world evaluation. For the GPU hardware platform, we utilize the CUTLASS toolkit, while the PyTorch toolkit is used for CPU and ARM hardware platforms.
\subsubsection{Quantization and Comparison Settings}
For a fair quantization experiment, EDA-DM configures the models and reconstruction the same way as Q-Diffusion~\cite{li2023q}.
we employ channel-wise quantization for weights and layer-wise quantization for activations, as it is a common setting.
In the reconstruction, we set the calibration samples to 1024 and the training batch to 32 for DDIMs and LDMs experiments. Due to time and memory source constraints, we adjust the reconstruction calibration samples to 256 and the training batch to 2 for Stable-Diffusion.
The notion ``WxAy'' is employed to represent the bit-widths of weights ``W''
and activations ``A''.
For the experimental comparison, we compare EDA-DM with the PTQ methods for diffusion models, including PTQ4DM~\cite{shang2023post}, Q-Diffusion~\cite{li2023q}, PTQD~\cite{he2023ptqd}, ADP-DM~\cite{wang2023towards}, TFMQ-DM~\cite{huang2024tfmq}, TAC-Diffusion~\cite{yao2024timestep}, and TCAQ-DM~\cite{huang2024tcaq}.

\subsubsection{Evaluation Metrics}
The evaluation metrics include FID, sFID, IS, CLIP Score (on ViT-g/14)~\cite{heusel2017fid,hessel2021clipscore,salimans2016IS}, and Aesthetic Score\footnote{\url{https://github.com/shunk031/simple-aesthetics-predictor}}. 
Following the common practice~\cite{shang2023post,li2023q}, the Stable-Diffusion generates 10,000 images, while all other models generate 50,000 images. 
Most of the existing methods employ hardware-unfriendly operations to improve accuracy, such as introducing additional overhead or a large number of quantization parameters. We use the ``Friendly'' metric to indicate the hardware-friendly nature of the method.
Additionally, we evaluate the model size and runtime before and after quantization to visualize the compression and acceleration effects of EDA-DM.
The speed up ratio is calculated by measuring the time taken to generate a single image on the RTX 3090.
We also assess the generation performance of the quantized models by visualizing random samples.

\begin{table}[h]\small %\small\footnotesize %\scriptsize %\tiny %\small
    \centering
    \begin{threeparttable}
    \setlength{\tabcolsep}{1.35mm}
    \caption{Quantization results of DDIM on CIFAR-10. }
    \vspace{-0.1cm}
    \begin{tabular}{cccccc}
    \toprule 
    Task & Method & Bit-width & Friendly & FID$\downarrow$ & IS$\uparrow$ \\
    % \midrule
    \cmidrule(r){1-6}
    \multirow{15}{1.55cm}{\centering \\CIFAR-10 \\32 $\times $ 32\\$ $\\DDIM\\steps = 100} & FP & W32A32 & - & 4.26 & 9.03 \\ 
    \cmidrule(r){2-6}
    \multirow{15}{*}{} & PTQ4DM\textsuperscript{$\star$} & W8A8 & \cmark & 4.39 & 9.25 \\
    \multirow{15}{*}{} & Q-Diffusion\textsuperscript{$\dagger$} & W8A8 & \cmark & 4.06 & 9.38 \\
    \multirow{15}{*}{} & TDQ\textsuperscript{$\dagger$} & W8A8 & \xmark & 5.99 & 8.85 \\
    \multirow{15}{*}{} & ADP-DM\textsuperscript{$\dagger$} & W8A8 & \xmark & 4.24 & 9.07 \\
    \multirow{15}{*}{} & TFMQ-DM\textsuperscript{$\dagger$} & W8A8 & \xmark & 4.24 & 9.07 \\
    \multirow{15}{*}{} & TAC-Diffusion\textsuperscript{$\dagger$} & W8A8 & \xmark & \textbf{3.68} & \textbf{9.49} \\
    \multirow{15}{*}{} & TCAQ-DM\textsuperscript{$\dagger$} & W8A8 & \xmark & 4.09 & 9.08 \\
    \multirow{15}{*}{} & EDA-DM & W8A8 & \cmark & 3.73 & 9.40 \\
    \cmidrule(r){2-6}
    \multirow{15}{*}{} & PTQ4DM\textsuperscript{$\star$} & W4A8 & \cmark & 5.31 & 9.24 \\
    \multirow{15}{*}{} & Q-Diffusion\textsuperscript{$\dagger$} & W4A8 & \cmark & 4.93 & 9.12 \\
    \multirow{15}{*}{} & TFMQ-DM\textsuperscript{$\dagger$} & W4A8 & \xmark & 4.78 & 9.13 \\
    \multirow{15}{*}{} & TAC-Diffusion\textsuperscript{$\dagger$} & W4A8 & \xmark & 4.89 & 9.15 \\
    \multirow{15}{*}{} & TCAQ-DM\textsuperscript{$\dagger$} & W4A8 & \xmark & 4.59 & 9.17 \\
    \multirow{15}{*}{} & EDA-DM & W4A8 & \cmark & \textbf{4.03} & \textbf{9.43} \\
    \bottomrule
\end{tabular}
 \begin{tablenotes}
    \scriptsize
    \item[] \textsuperscript{$\star$} denotes our implementation according to open-source codes.
    \item[] \textsuperscript{$\dagger$} represents results directly obtained by papers or re-running open-source codes.
 \end{tablenotes}
    \label{tab:uncondition}
    \end{threeparttable}
    \vspace{-0.3cm}
\end{table}
\begin{figure*}[!h]
    \centering
    \includegraphics[width=0.98\textwidth]{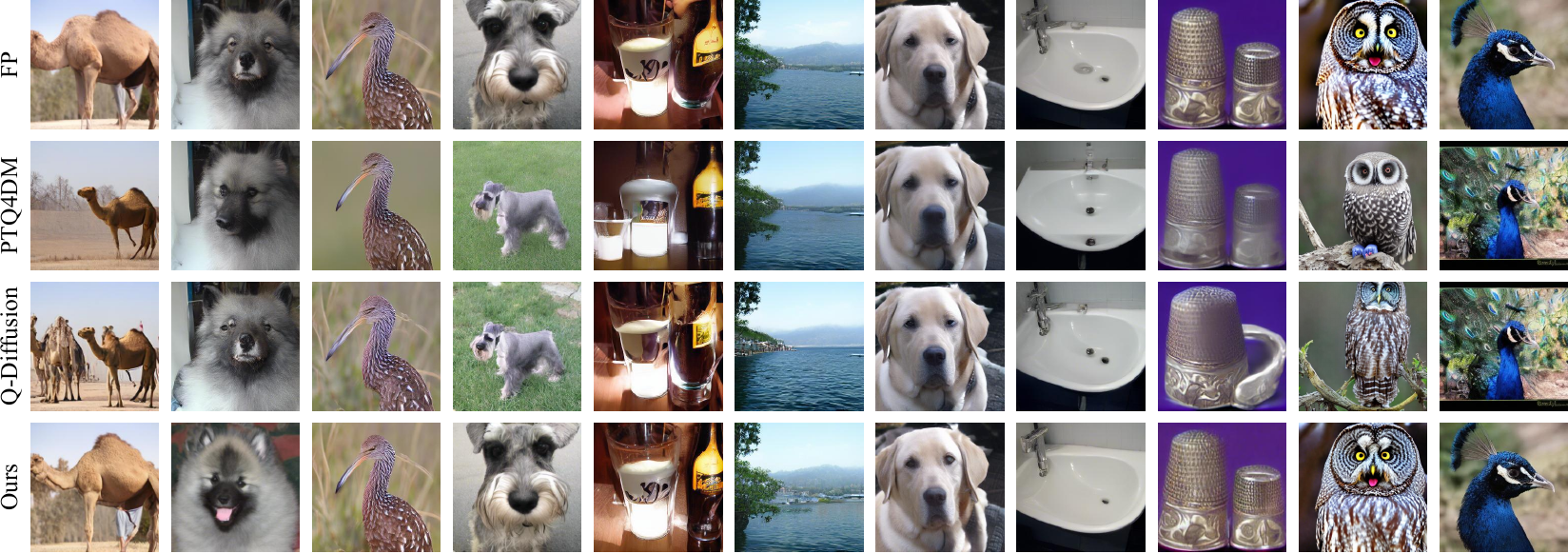}
    \vspace{-0.1cm}
    \caption{Random samples generated by LDM-4 model on ImageNet dataset at W4A8 precision.}
    \vspace{-0.3cm}
    \label{fig:imagenet}
\end{figure*}

\begin{table}[t]\small %\footnotesize %\scriptsize %\tiny %\small
    \centering
    \setlength{\tabcolsep}{1.15mm}
    \caption{Quantization results of LDM on LSUN.}
    \vspace{-0.1cm}
    \begin{tabular}{cccccc}
    \toprule 
    Task & Method & Bit-width & Fridndly & FID$\downarrow$ & sFID$\downarrow$ \\
    \cmidrule(r){1-6}
    \multirow{14}{1.6cm}{\centering \\LSUN \\ Bedroom \\256 $\times $ 256\\$ $\\LDM-4\\steps = 200\\eta = 1.0} & FP& W32A32 & - & 3.02 & 7.21 \\ 
    \cmidrule(r){2-6}
    \multirow{14}{*}{} & PTQ4DM\textsuperscript{$\star$} & W8A8 & \cmark & 4.18 & 9.59 \\
    \multirow{14}{*}{} & Q-Diffusion\textsuperscript{$\dagger$} & W8A8 & \cmark & 4.40 & 8.17 \\
    \multirow{14}{*}{} & PTQD\textsuperscript{$\dagger$} & W8A8 & \cmark & 3.75 & 9.89 \\
    \multirow{14}{*}{} & TFMQ-DM\textsuperscript{$\dagger$} & W8A8 & \xmark & \textbf{3.14} & \textbf{7.26} \\
    \multirow{14}{*}{} & TCAQ-DM\textsuperscript{$\dagger$} & W8A8 & \xmark & 3.21 & 7.59 \\
    \multirow{14}{*}{} & EDA-DM & W8A8 & \cmark & 3.46 & 7.50 \\
    \cmidrule(r){2-6}
    \multirow{14}{*}{} & PTQ4DM\textsuperscript{$\star$} & W4A8 & \cmark & 4.25 & 14.22 \\
    \multirow{14}{*}{} & Q-Diffusion\textsuperscript{$\dagger$} & W4A8 & \cmark & 5.32 & 16.82 \\
    \multirow{14}{*}{} & PTQD\textsuperscript{$\dagger$} & W4A8 & \cmark & 5.94 & 15.16 \\
    \multirow{14}{*}{} & TFMQ-DM\textsuperscript{$\dagger$} & W4A8 & \xmark & 3.68 & 7.65 \\
    \multirow{14}{*}{} & TAC-Diffusion\textsuperscript{$\dagger$} & W4A8 & \xmark & 4.94 & - \\
    \multirow{14}{*}{} & TCAQ-DM\textsuperscript{$\dagger$} & W4A8 & \xmark & 3.70 & 7.69 \\
    \multirow{14}{*}{} & EDA-DM & W4A8 & \cmark & \textbf{3.63} & \textbf{6.59} \\
    % \midrule
    \cmidrule(r){1-6}
    \multirow{13}{1.6cm}{\centering \\LSUN \\ Church \\256 $\times $ 256\\$ $\\LDM-8\\steps = 500\\eta = 0.0} & FP & 32/32 & - & 4.06 & 10.89 \\ 
    \cmidrule(r){2-6}
    \multirow{13}{*}{} & PTQ4DM\textsuperscript{$\star$} & W8A8 & \cmark & 3.98 & 13.48 \\
    \multirow{13}{*}{} & Q-Diffusion\textsuperscript{$\dagger$} & W8A8 & \cmark & \textbf{3.65} & 12.23 \\
    \multirow{13}{*}{} & PTQD\textsuperscript{$\star$} & W8A8 & \cmark & 4.13 & 13.89 \\
    \multirow{13}{*}{} & TFMQ-DM\textsuperscript{$\dagger$} & W8A8 & \xmark & 4.01 & 10.98 \\
    \multirow{13}{*}{} & TCAQ-DM\textsuperscript{$\dagger$} & W8A8 & \xmark & 4.05 & 10.82 \\
    \multirow{13}{*}{} & EDA-DM & W8A8 & \cmark & 3.83 & \textbf{10.75} \\
    \cmidrule(r){2-6}
    \multirow{13}{*}{} & PTQ4DM\textsuperscript{$\star$} & W4A8 & \cmark & 4.20 & 14.87 \\
    \multirow{13}{*}{} & Q-Diffusion\textsuperscript{$\dagger$} & W4A8 & \cmark & 4.12 & 13.94 \\
    \multirow{13}{*}{} & PTQD\textsuperscript{$\star$} & W4A8 & \cmark & 4.33 & 15.67 \\
    \multirow{13}{*}{} & TFMQ-DM\textsuperscript{$\dagger$} & W4A8 & \xmark & 4.14 & 11.46 \\
    \multirow{13}{*}{} & TCAQ-DM\textsuperscript{$\dagger$} & W4A8 & \xmark & 4.13 & 11.57 \\
    \multirow{13}{*}{} & EDA-DM & W4A8 & \cmark & \textbf{4.01} & \textbf{10.95} \\
    \bottomrule
\end{tabular}
    \label{tab:uncondition1}
    \vspace{-0.3cm}
\end{table}

\subsection{Main Results}
\subsubsection{Unconditional Image Generation}
The quantization results are reported in Table~\ref{tab:uncondition} and~\ref{tab:uncondition1}.
We focus on the performance of low-bit quantization to highlight the advantages of EDA-DM.
At W4A8 precision, EDA-DM achieves significant improvement with a notable 0.56 (4.03 vs. 4.59) FID score and 0.26 (9.43 vs. 9.17) IS score enhancement over TCAQ-DM on CIFAR-10.
It also significantly improves the quantization performance on LSUN-Bedroom and LSUN-Church, with sFID score reductions of 0.96 (6.59 vs. 7.65) and 0.51 (10.95 vs. 11.46) compared to TFMQ-DM, respectively.
Although TAC-Diffusion and TFMQ-DM achieve better performance on CIFAR-10 and LSUN-Bedroom at W8A8 precision, the introduction of hardware-unfriendly operations significantly reduce their acceleration performance. We further discuss this impact in the ablation study.
In contrast, EDA-DM not only maintains the hardware-friendly configuration, but it also outperforms even the full-precision models on CIFAR-10 and LSUN-Church at W4A8 precision.

\subsubsection{Class-Conditional Image Generation}
We conduct experiments on the ImageNet 256$\times$256 dataset, and the results are reported in Table~\ref{tab:condition}.
Compared to the state-of-the-art TCAQ-DM, our method improves the FID score by 0.13 and the sFID score by 1.90 at W4A8 precision. 
Besides, under hardware-friendly conditions, EDA-DM significantly improves the FID score by 0.56 (9.84 vs. 10.40) and the sFID score by 6.91 (5.77 vs. 12.68) compared to PTQD.
As shown in Fig.~\ref{fig:imagenet}, our method achieves superior generation quality compared to existing approaches and even outperforms the full-precision model.
\subsubsection{Text-Conditional Image Generation}
\begin{figure*}[!t]
    \centering
    \includegraphics[width=0.98\textwidth]{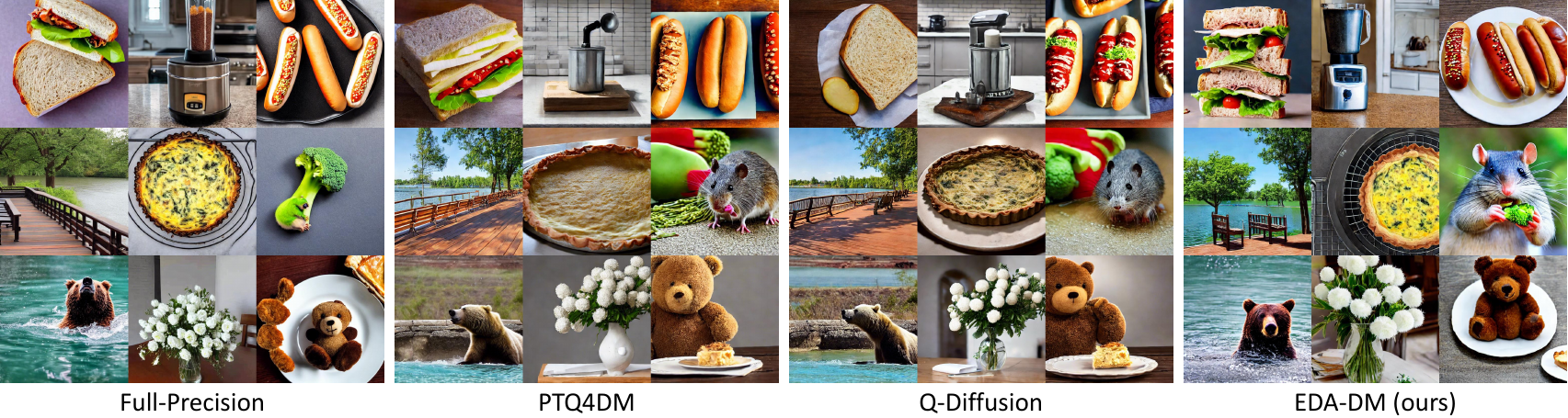}
    \vspace{-0.1cm}
    \caption{Random samples generated by Stable-Diffusion on COCO dataset at W4A8 precision.}
    \label{fig:coco}
    \vspace{-0.3cm}
\end{figure*}

In this experiment, we sample high-resolution images of 512$\times$512 pixels with Stable-Diffusion, which helps validate the robustness of our method for high-resolution and large models.
Compared to existing methods, EDA-DM achieves state-of-the-art performance at both W8A8 and W4A8 precision, as reported in Table~\ref{tab:uncondition1}.
Especially at W4A8 precision, EDA-DM narrows the CLIP score gap between quantized model and full-precision model to 0.17 and improves the FID score to 20.58.
This demonstrates that our method significantly preserves the semantic information and generation quality for text-to-image models.
We also visualize the generation quality of the quantized models in Fig.~\ref{fig:coco}.

\begin{table}[t]\footnotesize %\footnotesize %\scriptsize %\tiny %\small
    \centering
    \setlength{\tabcolsep}{1.15mm}
    \caption{Quantization results of class-guided image generation.}
    \vspace{-0.1cm}
    \begin{tabular}{ccccccc}
    \toprule
    Task & Method & Bit-width & Friendly & FID$\downarrow$ & sFID$\downarrow$ & IS$\uparrow$ \\
    \cmidrule(r){1-7}
    \multirow{13}{1.35cm}{\centering ImageNet \\256 $\times $ 256\\$ $\\LDM-4\\steps = 20\\eta = 0.0\\scale = 3.0} & FP & 32/32 & - & 11.69 & 7.67 & 364.73 \\ 
    \cmidrule(r){2-7}
    \multirow{13}{*}{} & PTQ4DM\textsuperscript{$\star$}  & W8A8 & \cmark & 11.57 & 9.82 & 350.24 \\
    \multirow{13}{*}{} & Q-Diffusion\textsuperscript{$\star$}  & W8A8 & \cmark & 11.59 & 9.87 & 347.43 \\
    \multirow{13}{*}{} & PTQD\textsuperscript{$\dagger$}  & W8A8 & \cmark & 11.94 & 8.03 & 350.26 \\
    \multirow{13}{*}{} & TFMQ-DM\textsuperscript{$\dagger$}  & W8A8 & \xmark & \textbf{10.50} & 7.96 & - \\
    \multirow{13}{*}{} & TCAQ-DM\textsuperscript{$\dagger$}  & W8A8 & \xmark & 10.58 & 7.54 & - \\
    \multirow{13}{*}{} & EDA-DM  & W8A8 & \cmark & 11.10 & \textbf{6.95} & \textbf{353.02} \\
    \cmidrule(r){2-7}
    \multirow{13}{*}{} & PTQ4DM\textsuperscript{$\star$}  & W4A8 & \cmark & 13.57 & 16.06 & 323.17 \\
    \multirow{13}{*}{} & Q-Diffusion\textsuperscript{$\star$}  & W4A8 & \cmark & 12.40 & 14.85 & 336.80 \\
    \multirow{13}{*}{} & PTQD\textsuperscript{$\dagger$}  & W4A8 & \cmark & 10.40 & 12.68 & 344.72 \\
    \multirow{13}{*}{} & TFMQ-DM\textsuperscript{$\dagger$}  & W4A8 & \xmark & 10.29 & 7.35 & - \\
    \multirow{13}{*}{} & TCAQ-DM\textsuperscript{$\dagger$}  & W4A8 & \xmark & 9.97 & 7.67 & - \\
    \multirow{13}{*}{} & EDA-DM  & W4A8 & \cmark & \textbf{9.84} & \textbf{5.77} & \textbf{348.75} \\
    \bottomrule
\end{tabular}
    \label{tab:condition}
    \vspace{-0.3cm}
\end{table}
\begin{table}[t]\footnotesize %\footnotesize %\scriptsize %\tiny %\small
    \centering
    \setlength{\tabcolsep}{1.15mm}
    \caption{Quantization results of text-guided image generation.}
    \vspace{-0.1cm}
    \begin{tabular}{ccccccc}
    \toprule
    Task & Method & Bit-width & Friendly & FID$\downarrow$ & sFID$\downarrow$ & CLIP$\uparrow$ \\
    \cmidrule(r){1-7}
    \multirow{9}{1.35cm}{\centering \\MS-COCO \\300 $\times $ 300 \\Stable-Diffusion\\steps = 50\\eta = 0.0\\scale = 7.5} & FP & W32A32 & - & 21.96 & 33.86 & 26.88 \\ 
    \cmidrule(r){2-7}
    \multirow{9}{*}{} & PTQ4DM\textsuperscript{$\star$} & W8A8 & \cmark & 20.48 & 33.08 & 26.79 \\
    \multirow{9}{*}{} & Q-Diffusion\textsuperscript{$\star$} & W8A8 & \cmark & 20.47 & 32.97 & 26.78 \\
    \multirow{9}{*}{} & TFMQ-DM\textsuperscript{$\star$} & W8A8 & \xmark & 20.17 & 32.57 & 26.78 \\
    \multirow{9}{*}{} & EDA-DM & W8A8 & \cmark & \textbf{19.97} & \textbf{32.22} & \textbf{26.83} \\
    \cmidrule(r){2-7}
    \multirow{9}{*}{} & PTQ4DM\textsuperscript{$\star$} & W4A8 & \cmark & 22.48 & 34.32 & 26.00 \\
    \multirow{9}{*}{} & Q-Diffusion\textsuperscript{$\star$} & W4A8 & \cmark & 21.96 & 33.81 & 26.29 \\
    \multirow{9}{*}{} & TFMQ-DM\textsuperscript{$\star$} & W4A8 & \xmark & 21.94 & \textbf{32.84} & 26.56 \\
    \multirow{9}{*}{} & EDA-DM & W4A8 & \cmark & \textbf{20.58} & 33.08 & \textbf{26.71} \\
    \bottomrule
\end{tabular}
    \label{tab:condition1}
    \vspace{-0.3cm}
\end{table}
\begin{table}[!t]\small
    \centering
    \caption{Aesthetic Score of quantized models at W4A8 precision.}
    \vspace{-0.1cm}
    \setlength{\tabcolsep}{1.35mm}
    \begin{tabular}{cccc}
        \toprule
        Method & LSUN-Bedroom & LSUN-Church & DrawBench \\
        \cmidrule(r){1-4}
        FP & 5.91 & 5.88 & 5.80 \\
        \cmidrule(r){1-4}
        Q-Diffusion & 5.82 & 5.75 & 5.60 \\
        TFMQ-DM & 5.83 & 5.77 & 5.60 \\
        EDA-DM & \bf 5.87 & \bf 5.80 & \bf 5.66 \\
        \bottomrule
    \end{tabular}
    \label{tab:aes}
    \vspace{-0.3cm}
\end{table}
\subsubsection{Human Preference Evaluation}
\begin{figure*}[!t]
    \centering
    \includegraphics[width=0.98\textwidth]{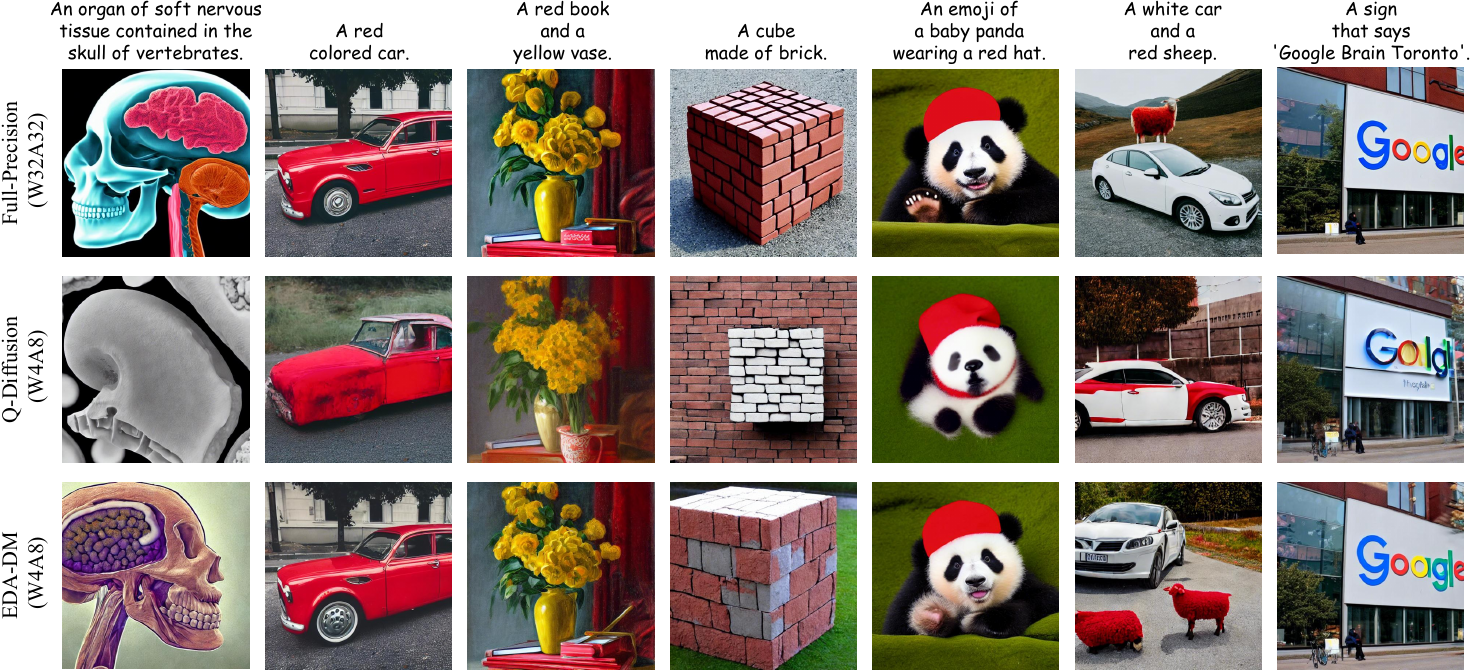}
    \vspace{-0.1cm}
    \caption{Human performance evaluation for Stable-Diffusion.}
    \label{fig:human}
    \vspace{-0.3cm}
\end{figure*}

Considering that automated metrics do not fully represent the quality of generation, we further evaluate human preferences by assessing Aesthetic Score~$\uparrow$ and visualizing random samples.
As reported in Table~\ref{tab:aes}, the quantized model with our method enable to generate images that are more aesthetically pleasing to humans.
We use the convincing DrawBench benchmark to evaluate the quantized Stable-Diffusion.
As shown in Fig.~\ref{fig:human}, due to the low-bit quantization, the quantized model cannot generate images exactly the same as the full-precision model.
However, when compared to other methods, EDA-DM significantly preserves the semantic information and generation quality.

\subsection{Analysis}\label{sec:4.4}
\subsubsection{Ablation Study}
We conduct experiments for LDM-4 on ImageNet to showcase the effect of different components of our method.
As shown in Table~\ref{tab:ablation1}, the baseline employs random sampling calibration combined with block-wise reconstruction.
By introducing TDAC and FBR, the FID score is improved to 10.75 and 10.55, respectively.
Furthermore, using the two components of our method, the FID score can be significantly improved to 9.84.

To demonstrate the advantages of TDAC and FBR in detail, we replaced the sampling strategy and reconstruction method of the DDIM baseline on CIFAR-10 with different approaches.
As reported in Table~\ref{tab:ablation}, TDAC outperforms the compared sampling strategies across different numbers of calibration samples, demonstrating its robustness to the size of the calibration.
Besides, FBR effectively addresses the issues of overfitting and underfitting in reconstruction, surpassing existing reconstruction methods.

\begin{table}[t]\small %\footnotesize %\scriptsize %\tiny %\small
    \centering
    \caption{The effect of different components proposed in the paper. }
    \vspace{-0.1cm}
    \label{tab:ablation1}
    \setlength{\tabcolsep}{1.5mm}
    \begin{tabular}{ccccc}
        \toprule 
        \bf Method & \bf Bit-width & \bf FID $\downarrow$ & \bf sFID $\downarrow$ & \bf IS $\uparrow$ \\ 
        \cmidrule(r){1-5}
        baseline & W4A8 & 16.23 & 9.78 & 324.96 \\ 
        +TDAC & W4A8 & 10.75 & 9.45 & 337.81 \\ 
        +FBR & W4A8 & 10.55 & 6.35 & 354.16 \\
        +TDAC+FBR & W4A8 & 9.84 & 5.77 & 348.75 \\
        \bottomrule
    \end{tabular}
    \vspace{-0.3cm}
\end{table}
\begin{table}[!t]\small %\footnotesize %\scriptsize %\tiny %\small
    \centering
    \caption{Advantages of our methods. Here, DNTC, TSC, and TSS are sampling strategies in PTQ4DM, Q-Diffusion, and APQ-DM, respectively. TIAR is the reconstruction method in TFMQ-DM.}
    \vspace{-0.1cm}
    \label{tab:ablation}
    \setlength{\tabcolsep}{3.0mm}
    \begin{tabular}{ccc|cc}
    \toprule
    \multicolumn{1}{c}{\multirow{3}{1.6cm}{\centering \\Method}} & \multicolumn{4}{c}{Calibration}\\ 
    \cmidrule(r){2-5}
    \multicolumn{1}{c}{\multirow{3}{1.6cm}{}} & \multicolumn{2}{c|}{1024} & \multicolumn{2}{c}{5120}\\
    \cmidrule(r){2-5}
    \multicolumn{1}{c}{\multirow{3}{1.6cm}{}} & FID$\downarrow$ & IS$\uparrow$ & FID$\downarrow$ & IS$\uparrow$\\ 
    \cmidrule(r){1-5}
    FP & 4.26 & 9.03 & 4.26 & 9.03 \\
    \cmidrule(r){1-5}
    NDTC & 5.31 & 9.24 & 6.48 & 9.10 \\
    TSC & 4.55 & 9.36 & 4.93 & 9.12 \\
    TSS & 5.76 & 9.16 & 6.07 & 9.11 \\
    TDAC (ours) & \textbf{4.42} & \textbf{9.38} & \textbf{4.40} & \textbf{9.45} \\
    \cmidrule(r){1-5}
    Layer-wise & 4.70 & 9.36 & 5.04 & 9.43 \\
    Block-wise & 4.55 & 9.36 & 4.93 & 9.12 \\
    TIAR & 4.40 & 9.40 & 4.56 & 9.24 \\
    FBR (ours) & \textbf{4.21} & \textbf{9.48} & \textbf{4.29} & \textbf{9.47}
    \\
    \bottomrule
\end{tabular}
    \vspace{-0.45cm}
\end{table}

\subsubsection{Robustness of Hyperparameter}
\begin{figure}[!b]
    \vspace{-0.3cm}
    \centering
    \includegraphics[width=0.95\linewidth]{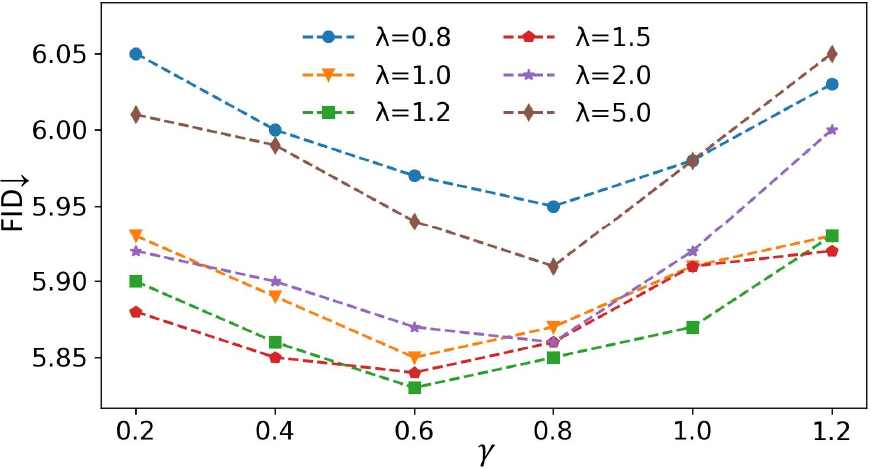}
    \vspace{-0.3cm}
    \caption{The performance w.r.t. different hyperparameters $\lambda$ and $\gamma$.}
    \label{fig:hyper}
\end{figure}

Our method involves two hyperparameters: $\lambda$ balancing the two scores for TDAC, and $\gamma$ coordinating the losses of block and layers for FBR.
We use the quantized DDIM on CIFAR-10 to generate 10,000 images for evaluation.
As shown in Fig.~\ref{fig:hyper}, the results obtained with a wide range of $\lambda$ and $\gamma$ outperform the works PTQ4DM (FID 6.91) and Q-Diffusion (FID 6.54).
This demonstrates that our method is robust to hyperparameters and easily migrates to other quantization tasks. 
The hyperparameters for other tasks are reported in Table~\ref{tab:hyper}.
Given the small size of the calibration for Stable-Diffusion, the $\lambda $ is set to 5.0.

\begin{table}[!t]\small %\footnotesize %\scriptsize %\tiny %\small
    \centering
    \caption{Hyperparameters for all experiments.}
    \vspace{-0.1cm}
    \setlength{\tabcolsep}{2.0mm}
    \begin{tabular}{cccc}
        \toprule
        Experiments & Calibration & $\lambda $ & $\gamma $ \\
        \cmidrule(r){1-4}
        DDIM on CIFAR-10 & 1024 & 1.2 & 0.8 \\
        LDM-4 on LSUN-Bedroom & 1024 & 1.0 & 1.0 \\
        LDM-8 on LSUN-Church & 1024 & 1.0 & 1.0 \\
        LDM-4 on ImageNet & 1024 & 1.2 & 0.8 \\
        Stable-Diffusion on COCO & 256 & 5.0 & 0.8 \\
        \bottomrule
    \end{tabular}
    \label{tab:hyper}
    \vspace{-0.3cm}
\end{table}

\subsubsection{Effectiveness of Two Scores}\label{sec:score}
\begin{figure}[!h]
    \centering
    \includegraphics[width=0.99\linewidth]{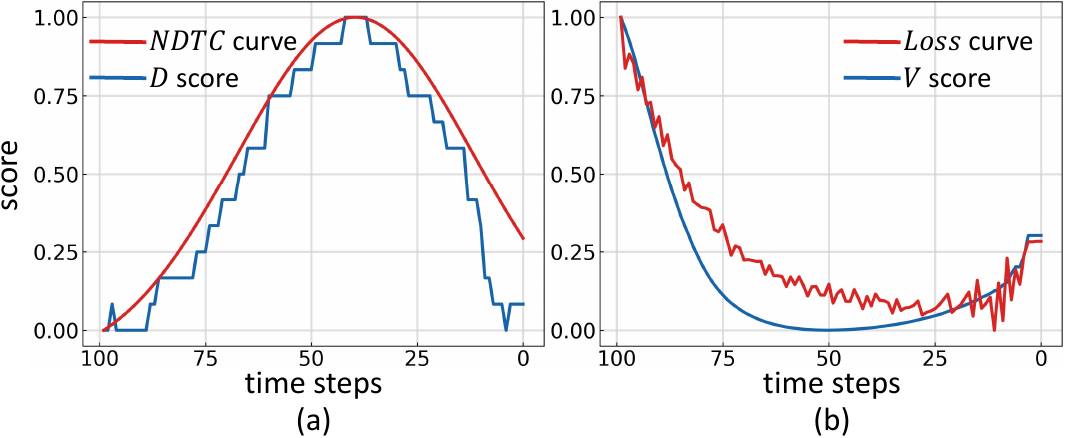}
    \vspace{-0.7cm}
    \caption{Effectiveness of two scores. Here, \textit{Loss} curve is the MSE error of the network output before and after quantization at different time steps. Data from DDIM on CIFAR-10 at W4A8 precision.}
    \label{fig:score}
    \vspace{-0.1cm}
\end{figure}

It is likely not feasible to demonstrate the effectiveness of the two scores separately through the performance of quantized models, since samples matched the density distribution and hard samples are both important for quantization.
So we demonstrate their effectiveness through the existing arguments and intuitive experiment.
As shown in Fig.~\ref{fig:score}, the $D$ score is consistent with the \emph{NDTC} curve~\cite{shang2023post}, which is designed through extensive experiments to fit the overall samples.
This shows that $D$ can reasonably represent the distribution of the overall sample.
The $V$ score is consistent with the quantization \emph{Loss} curve, which demonstrates that $V$ can effectively find hard samples through the diversity of feature maps.

\subsubsection{Robustness of Samplers and Steps}\label{app:sampler}
We perform ablation experiments to check the robustness of EDA-DM to samplers and steps.
As reported in Table~\ref{tab:sampler}, EDA-DM outperforms existing methods across different samplers and steps.

\begin{table}[t]\footnotesize %\footnotesize %\scriptsize %\tiny %\small
    \centering
    \caption{The performance of EDA-DM on different samplers and steps.}
    \vspace{-0.1cm}
    \setlength{\tabcolsep}{1.5mm}
    \begin{tabular}{cccccc}
    \toprule 
    Task & Method & Bit-width & FID$\downarrow$ & sFID$\downarrow$ & IS$\uparrow$ \\
    % \midrule
    \cmidrule(r){1-6}
    \multirow{7}{2.0cm}{\centering \\LDM-4\\DDIM~\cite{song2020denoising} \\time steps = 20} & FP & W32A32 & 11.69 & 7.67 & 364.73 \\ 
    \cmidrule(r){2-6}
    \multirow{7}{*}{} & Q-Diffusion\textsuperscript & W8A8 & 11.59 & 9.87 & 347.43 \\
    \multirow{7}{*}{} & PTQD\textsuperscript & W8A8 & 11.94 & 8.03 & 350.26 \\
    \multirow{7}{*}{} & Ours & W8A8 & \textbf{11.10} & \textbf{6.95} & \textbf{353.02} \\
    \cmidrule(r){2-6}
    \multirow{7}{*}{} & Q-Diffusion\textsuperscript & W4A8 & 12.40 & 14.85 & 336.80 \\
    \multirow{7}{*}{} & PTQD\textsuperscript & W4A8 & 10.40 & 12.68 & 344.72 \\
    \multirow{7}{*}{} & Ours & W4A8 & \textbf{9.84} & \textbf{5.77} & \textbf{348.75} \\
    \cmidrule(r){1-6}
    \multirow{7}{2.0cm}{\centering \\LDM-4\\ PLMS~\cite{liu2022pseudo} \\time steps = 20} & FP& 32/32& 11.71 & 6.08 & 379.19 \\ 
    \cmidrule(r){2-6}
    \multirow{7}{*}{} & Q-Diffusion\textsuperscript & W8A8 & 11.25 & 7.75 & 360.49 \\
    \multirow{7}{*}{} & PTQD\textsuperscript & W8A8 & 11.05 & \textbf{7.42} & 361.13 \\
    \multirow{7}{*}{} & Ours & W8A8 & \textbf{10.91} & 7.61 & \textbf{363.53} \\
    \cmidrule(r){2-6}
    \multirow{7}{*}{} & Q-Diffusion\textsuperscript & W4A8 & 11.27 & 5.74 & 358.13 \\
    \multirow{7}{*}{} & PTQD\textsuperscript & W4A8 & 10.84 & 5.96 & 357.66 \\
    \multirow{7}{*}{} & Ours & W4A8 & \textbf{10.74} & \textbf{5.68} & \textbf{359.60} \\
    \cmidrule(r){1-6}
    \multirow{7}{2.0cm}{\centering \\LDM-4\\ DPM-Solver~\cite{lu2022dpm} \\time steps = 20} & FP & 32/32 & 11.44 & 6.85 & 373.12 \\ 
    \cmidrule(r){2-6}
    \multirow{7}{*}{} & Q-Diffusion\textsuperscript & W8A8 & 10.78 & 7.15 & 342.64 \\
    \multirow{7}{*}{} & PTQD\textsuperscript & W8A8 & 10.66 & 6.73 & 348.22 \\
    \multirow{7}{*}{} & Ours & W8A8 & \textbf{10.58} & \textbf{6.55} & \textbf{352.51} \\
    \cmidrule(r){2-6}
    \multirow{7}{*}{} & Q-Diffusion\textsuperscript & W4A8 & 9.36 & 6.86 & 351.00 \\
    \multirow{7}{*}{} & PTQD\textsuperscript & W4A8 & 8.88 & 6.73 & 354.94 \\
    \multirow{7}{*}{} & Ours & W4A8 & \textbf{8.52} & \textbf{6.45} & \textbf{360.85} \\
    \cmidrule(r){1-6}
    \multirow{5}{2.0cm}{\centering \\LDM-4\\ DDIM~\cite{song2020denoising} \\time steps = 250} & FP & 32/32 & 3.37 & 5.14 & 204.56 \\ 
    \cmidrule(r){2-6}
    \multirow{5}{*}{} & Q-Diffusion\textsuperscript & W8A8 & 5.21 & 6.15 & 175.31 \\
    \multirow{5}{*}{} & Ours & W8A8 & \textbf{4.13} & \textbf{5.37} & \textbf{186.78} \\
    \cmidrule(r){2-6}
    \multirow{5}{*}{} & Q-Diffusion\textsuperscript & W4A8 & 6.36 & 6.89 & 170.21 \\
    \multirow{5}{*}{} & Ours & W4A8 & \textbf{4.79} & \textbf{5.68} & \textbf{176.43} \\
    \bottomrule
\end{tabular}

    \label{tab:sampler}
    \vspace{-0.3cm}
\end{table}

\subsubsection{Impact of Hardware-Unfriendly Settings}
Some methods improve model accuracy by introducing the hardware-unfriendly quantization settings.
For instance, APQ-DM introduces 8$\times$ quantization parameters and additional computation overhead to dynamically calculate the quantized values.
As reported in Table~\ref{tab:apq}, compared to standard quantization method (PTQ4DM), the additional computations make APQ-DM require more bit operations (Bops), leading to a reduced speedup ratio.
In addition, the increased quantization parameters reduce the model's compression efficiency. More importantly, it requires 8$\times$ the hardware resources for support.
As a result, these hardware-unfriendly settings compromise deployment efficiency.
In contrast, EDA-DM maintains the hardware-friendly settings and significantly improves performance.

\begin{table}[!t]\footnotesize %\footnotesize %\scriptsize %\tiny %\small
    \centering
    \caption{Deployment efficiency of different methods at 8-bit.}
    \vspace{-0.1cm}
    \setlength{\tabcolsep}{1.5mm}
    \begin{tabular}{c|ccccc}
        \toprule
        Method & Bops & Speedup & Model Size & Hardware & FID$\downarrow$ \\
        \midrule
        PTQ4DM & 402 G & 2.22$\times$ & 35.9 MB & 1$\times$ & 4.39 \\
        APQ-DM & 436 G & 1.76$\times$ & 42.7 MB & 8$\times$ & 4.24 \\
        EDA-DM & 402 G & 2.22$\times$ & 35.9 MB & 1$\times$ & 3.73 \\
        \bottomrule
    \end{tabular}
    \label{tab:apq}
    \vspace{-0.3cm}
\end{table}
\subsubsection{Deployment of Quantized Diffusion Models}
\begin{figure}[h]
    \centering
    \includegraphics[width=1.0\linewidth]{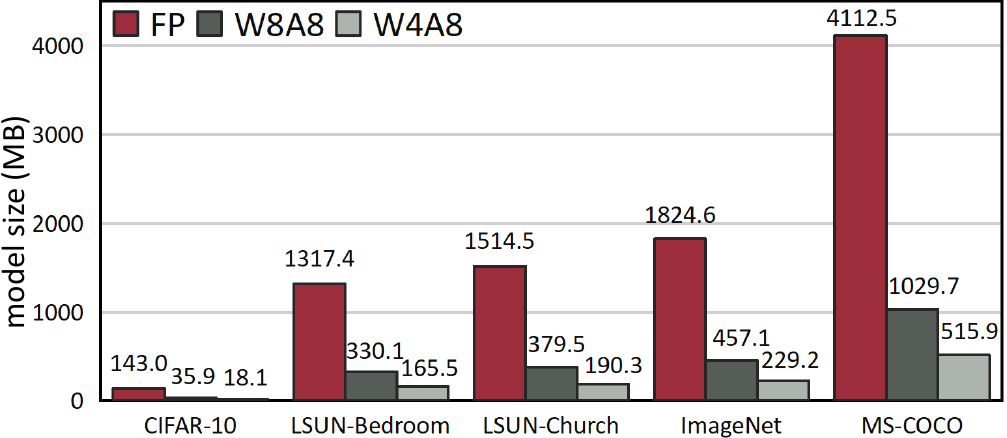}
    \vspace{-0.7cm}
    \caption{Model sizes of quantized diffusion models.}
    \label{fig:size}
    \vspace{-0.1cm}
\end{figure}

\begin{figure}[h]
    \centering
    \includegraphics[width=1.0\linewidth]{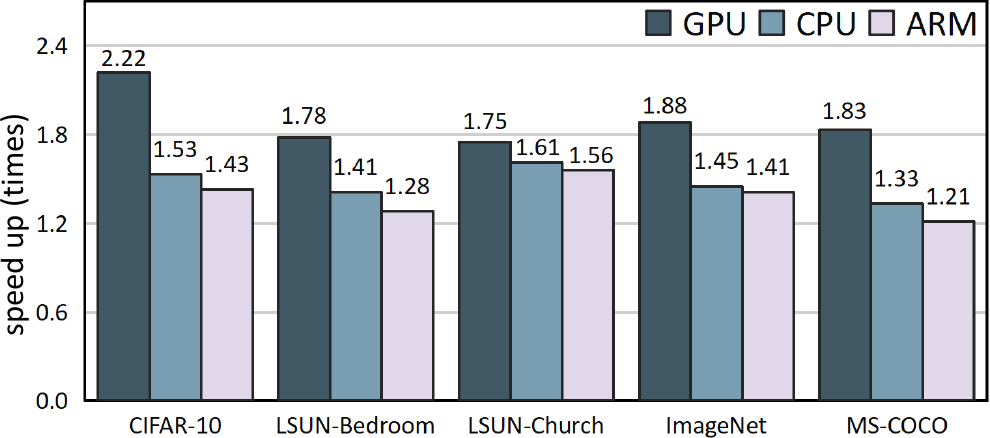}
    \vspace{-0.7cm}
    \caption{Speedup ratio on various hardware platforms.}
    \label{fig:speed}
    \vspace{-0.1cm}
\end{figure}

We deploy the 8-bit quantized models across various hardware platforms (GPU, CPU, ARM).
As shown in Fig.~\ref{fig:size} and~\ref{fig:speed}, EDA-DM compresses Stable-Diffusion from 4112.5 MB to 515.9 MB and achieves a 1.83$\times $ speedup on the GPU, significantly facilitating the real-world applications of text-to-image models.
We also present more intuitive acceleration and compression results in Table~\ref{tab:deploy}.

\begin{table}[!h]\footnotesize %\footnotesize %\scriptsize %\tiny %\small
    \centering
    \caption{Real-world evaluation of LDM-4 on ImageNet.}
    \vspace{-0.1cm}
    \setlength{\tabcolsep}{1.3mm}
    \begin{tabular}{c|ccccc}
        \toprule
        Method & Bit-width & Model Size & Runtime & Memory(GPU) & Speedup \\
        \midrule
        LDM-4 & W32A32 & 1824.6 MB & 360 ms & 10320 MB & 1.00$\times$ \\
        Ours & W8A8 & 457.1 MB & 191 ms & 6903 MB & 1.88$\times$ \\
        \bottomrule
    \end{tabular}
    \label{tab:deploy}
    \vspace{-0.3cm}
\end{table}
\section{Conclusion and Future Works}
\subsection{Conclusion}
In this paper, we identify the challenges of PTQ for diffusion models as the two levels of mismatch.
Based on the insight, we propose EDA-DM, a novel PTQ method to address these issues.
Specifically, at the calibration sample level, TDAC select samples based on feature maps in the temporal network to align the calibration samples with overall samples; at the reconstruction output level, FBR optimizes the loss of block-wise reconstruction with the losses of layers, aligning the quantized models and full-precision models at different network granularity.
Extensive experiments show that EDA-DM significantly outperforms existing methods across various models and different datasets.
Our method maintains deployment efficiency through the hardware-friendly settings, and we deploy the quantized models across different hardware platforms.
Furthermore, sufficient ablation studies demonstrate that EDA-DM is robust to samplers, steps, and hyperparameters.
This work provides a standardized and efficient quantization method to facilitate the real-world applications of diffusion models.
\subsection{Limitation and Future Works}
Although EDA-DM achieves remarkable performance at W8A8 and W4A8 precision, it experiences a certain degree of performance degradation at W4A4 precision. Moreover, EDA-DM has so far only been applied to diffusion models with a UNet framework, leaving models with other frameworks unexplored.
In the future, we will further refine EDA-DM to improve its compatibility with W4A4 precision and extend its application to diffusion models with alternative frameworks, such as the DiT~\cite{Peebles2022DiT} framework.

{\appendix
\subsection{Proof of Quantization Error} \label{sec:app}
Based on the Eq.~\ref{eq:quant}, the quantization-dequantization process of a activation element $x$ can be represented as:
\begin{align}
    Quant&: \bar{x} =clip \left( \left\lfloor\frac{x}{s}\right\rceil+z\right) \\
    DeQuant&: \hat{x}=s \cdot  \left( \bar{x}-z \right) \approx x
\end{align}

For the clarity of the derivation in Sec.~\ref{sec:3.3}, we express the introduction of quantization error to $x$ as $\hat{x}=x\cdot \left ( 1+u\left ( x \right )  \right ) $, where $u$ can be defined as:
\begin{equation}
    \begin{aligned}
        u & =\frac{\hat{x}}{x}-1 \\
        & =\frac{(\bar{x}-z) \cdot s}{(\bar{x}-z+c) \cdot s}-1 \\
        & =\frac{\bar{x}-z}{\bar{x}-z+c}-1 \\
        & =\frac{-c}{\bar{x}-z+c}
    \end{aligned}
\end{equation}
here, $c$ represents the quantization error, which is affected by bit-width and rounding error.
}

\bibliographystyle{IEEEtran}
\bibliography{egbib}

\begin{thebibliography}{10}
\providecommand{\url}[1]{#1}
\csname url@samestyle\endcsname
\providecommand{\newblock}{\relax}
\providecommand{\bibinfo}[2]{#2}
\providecommand{\BIBentrySTDinterwordspacing}{\spaceskip=0pt\relax}
\providecommand{\BIBentryALTinterwordstretchfactor}{4}
\providecommand{\BIBentryALTinterwordspacing}{\spaceskip=\fontdimen2\font plus
\BIBentryALTinterwordstretchfactor\fontdimen3\font minus \fontdimen4\font\relax}
\providecommand{\BIBforeignlanguage}[2]{{%
\expandafter\ifx\csname l@#1\endcsname\relax
\typeout{** WARNING: IEEEtran.bst: No hyphenation pattern has been}%
\typeout{** loaded for the language `#1'. Using the pattern for}%
\typeout{** the default language instead.}%
\else
\language=\csname l@#1\endcsname
\fi
#2}}
\providecommand{\BIBdecl}{\relax}
\BIBdecl

\bibitem{sohl2015deep}
J.~Sohl-Dickstein, E.~Weiss, N.~Maheswaranathan, and S.~Ganguli, ``Deep unsupervised learning using nonequilibrium thermodynamics,'' in \emph{International conference on machine learning}.\hskip 1em plus 0.5em minus 0.4em\relax PMLR, 2015, pp. 2256--2265.

\bibitem{ho2020denoising}
J.~Ho, A.~Jain, and P.~Abbeel, ``Denoising diffusion probabilistic models,'' \emph{Advances in neural information processing systems}, vol.~33, pp. 6840--6851, 2020.

\bibitem{niu2020permutation}
C.~Niu, Y.~Song, J.~Song, S.~Zhao, A.~Grover, and S.~Ermon, ``Permutation invariant graph generation via score-based generative modeling,'' in \emph{International Conference on Artificial Intelligence and Statistics}.\hskip 1em plus 0.5em minus 0.4em\relax PMLR, 2020, pp. 4474--4484.

\bibitem{ozbey2023unsupervised}
M.~Ozbey, O.~Dalmaz, S.~U. Dar, H.~A. Bedel, S.~Ozturk, A.~Gungor, and T.~Cukur, ``Unsupervised medical image translation with adversarial diffusion models.'' \emph{IEEE Transactions on Medical Imaging}, 2023.

\bibitem{10685062}
Y.~Xing, L.~Qu, S.~Zhang, K.~Zhang, Y.~Zhang, and L.~Bruzzone, ``Crossdiff: Exploring self-supervisedrepresentation of pansharpening via cross-predictive diffusion model,'' \emph{IEEE Transactions on Image Processing}, vol.~33, pp. 5496--5509, 2024.

\bibitem{lugmayr2022repaint}
A.~Lugmayr, M.~Danelljan, A.~Romero, F.~Yu, R.~Timofte, and L.~Van~Gool, ``Repaint: Inpainting using denoising diffusion probabilistic models,'' in \emph{Proceedings of the IEEE/CVF Conference on Computer Vision and Pattern Recognition}, 2022, pp. 11\,461--11\,471.

\bibitem{10566611}
X.~Gao, Y.~Yang, Y.~Wu, S.~Du, and G.-J. Qi, ``Multi-condition latent diffusion network for scene-aware neural human motion prediction,'' \emph{IEEE Transactions on Image Processing}, vol.~33, pp. 3907--3920, 2024.

\bibitem{10494219}
S.~Welker, H.~N. Chapman, and T.~Gerkmann, ``Driftrec: Adapting diffusion models to blind jpeg restoration,'' \emph{IEEE Transactions on Image Processing}, vol.~33, pp. 2795--2807, 2024.

\bibitem{saharia2022photorealistic}
C.~Saharia, W.~Chan, S.~Saxena, L.~Li, J.~Whang, E.~L. Denton, K.~Ghasemipour, R.~Gontijo~Lopes, B.~Karagol~Ayan, T.~Salimans \emph{et~al.}, ``Photorealistic text-to-image diffusion models with deep language understanding,'' \emph{Advances in neural information processing systems}, vol.~35, pp. 36\,479--36\,494, 2022.

\bibitem{luo2023latent}
S.~Luo, Y.~Tan, L.~Huang, J.~Li, and H.~Zhao, ``Latent consistency models: Synthesizing high-resolution images with few-step inference,'' \emph{arXiv preprint arXiv:2310.04378}, 2023.

\bibitem{yang2024cogvideox}
Z.~Yang, J.~Teng, W.~Zheng, M.~Ding, S.~Huang, J.~Xu, Y.~Yang, W.~Hong, X.~Zhang, G.~Feng \emph{et~al.}, ``Cogvideox: Text-to-video diffusion models with an expert transformer,'' \emph{arXiv preprint arXiv:2408.06072}, 2024.

\bibitem{khachatryan2023text2video}
L.~Khachatryan, A.~Movsisyan, V.~Tadevosyan, R.~Henschel, Z.~Wang, S.~Navasardyan, and H.~Shi, ``Text2video-zero: Text-to-image diffusion models are zero-shot video generators,'' in \emph{Proceedings of the IEEE/CVF International Conference on Computer Vision}, 2023, pp. 15\,954--15\,964.

\bibitem{nichol2021improved}
A.~Q. Nichol and P.~Dhariwal, ``Improved denoising diffusion probabilistic models,'' in \emph{International Conference on Machine Learning}.\hskip 1em plus 0.5em minus 0.4em\relax PMLR, 2021, pp. 8162--8171.

\bibitem{song2020denoising}
J.~Song, C.~Meng, and S.~Ermon, ``Denoising diffusion implicit models,'' \emph{arXiv preprint arXiv:2010.02502}, 2020.

\bibitem{10738304}
C.~Xu, J.~Yan, M.~Yang, and C.~Deng, ``Rethinking noise sampling in class-imbalanced diffusion models,'' \emph{IEEE Transactions on Image Processing}, vol.~33, pp. 6298--6308, 2024.

\bibitem{lu2022dpm}
C.~Lu, Y.~Zhou, F.~Bao, J.~Chen, C.~Li, and J.~Zhu, ``Dpm-solver++: Fast solver for guided sampling of diffusion probabilistic models,'' \emph{arXiv preprint arXiv:2211.01095}, 2022.

\bibitem{liu2022pseudo}
L.~Liu, Y.~Ren, Z.~Lin, and Z.~Zhao, ``Pseudo numerical methods for diffusion models on manifolds,'' \emph{arXiv preprint arXiv:2202.09778}, 2022.

\bibitem{rombach2022high}
R.~Rombach, A.~Blattmann, D.~Lorenz, P.~Esser, and B.~Ommer, ``High-resolution image synthesis with latent diffusion models,'' in \emph{Proceedings of the IEEE/CVF conference on computer vision and pattern recognition}, 2022, pp. 10\,684--10\,695.

\bibitem{gong2019differentiable}
R.~Gong, X.~Liu, S.~Jiang, T.~Li, P.~Hu, J.~Lin, F.~Yu, and J.~Yan, ``Differentiable soft quantization: Bridging full-precision and low-bit neural networks,'' in \emph{Proceedings of the IEEE/CVF international conference on computer vision}, 2019, pp. 4852--4861.

\bibitem{nagel2022overcoming}
M.~Nagel, M.~Fournarakis, Y.~Bondarenko, and T.~Blankevoort, ``Overcoming oscillations in quantization-aware training,'' in \emph{International Conference on Machine Learning}.\hskip 1em plus 0.5em minus 0.4em\relax PMLR, 2022, pp. 16\,318--16\,330.

\bibitem{li2023repq}
Z.~Li, J.~Xiao, L.~Yang, and Q.~Gu, ``Repq-vit: Scale reparameterization for post-training quantization of vision transformers,'' in \emph{Proceedings of the IEEE/CVF International Conference on Computer Vision}, 2023, pp. 17\,227--17\,236.

\bibitem{nagel2020up}
M.~Nagel, R.~A. Amjad, M.~Van~Baalen, C.~Louizos, and T.~Blankevoort, ``Up or down? adaptive rounding for post-training quantization,'' in \emph{International Conference on Machine Learning}.\hskip 1em plus 0.5em minus 0.4em\relax PMLR, 2020, pp. 7197--7206.

\bibitem{so2024temporal}
J.~So, J.~Lee, D.~Ahn, H.~Kim, and E.~Park, ``Temporal dynamic quantization for diffusion models,'' \emph{Advances in Neural Information Processing Systems}, vol.~36, 2024.

\bibitem{krizhevsky2009learning}
A.~Krizhevsky, G.~Hinton \emph{et~al.}, ``Learning multiple layers of features from tiny images,'' 2009.

\bibitem{he2023efficientdm}
Y.~He, J.~Liu, W.~Wu, H.~Zhou, and B.~Zhuang, ``Efficientdm: Efficient quantization-aware fine-tuning of low-bit diffusion models,'' \emph{arXiv preprint arXiv:2310.03270}, 2023.

\bibitem{hu2021lora}
E.~J. Hu, Y.~Shen, P.~Wallis, Z.~Allen-Zhu, Y.~Li, S.~Wang, L.~Wang, and W.~Chen, ``Lora: Low-rank adaptation of large language models,'' \emph{arXiv preprint arXiv:2106.09685}, 2021.

\bibitem{shang2023post}
Y.~Shang, Z.~Yuan, B.~Xie, B.~Wu, and Y.~Yan, ``Post-training quantization on diffusion models,'' in \emph{Proceedings of the IEEE/CVF Conference on Computer Vision and Pattern Recognition}, 2023, pp. 1972--1981.

\bibitem{li2023q}
X.~Li, Y.~Liu, L.~Lian, H.~Yang, Z.~Dong, D.~Kang, S.~Zhang, and K.~Keutzer, ``Q-diffusion: Quantizing diffusion models,'' in \emph{Proceedings of the IEEE/CVF International Conference on Computer Vision}, 2023, pp. 17\,535--17\,545.

\bibitem{wang2023towards}
C.~Wang, Z.~Wang, X.~Xu, Y.~Tang, J.~Zhou, and J.~Lu, ``Towards accurate data-free quantization for diffusion models,'' \emph{arXiv preprint arXiv:2305.18723}, 2023.

\bibitem{li2021brecq}
Y.~Li, R.~Gong, X.~Tan, Y.~Yang, P.~Hu, Q.~Zhang, F.~Yu, W.~Wang, and S.~Gu, ``Brecq: Pushing the limit of post-training quantization by block reconstruction,'' \emph{arXiv preprint arXiv:2102.05426}, 2021.

\bibitem{hubara2020improving}
I.~Hubara, Y.~Nahshan, Y.~Hanani, R.~Banner, and D.~Soudry, ``Improving post training neural quantization: Layer-wise calibration and integer programming,'' \emph{arXiv preprint arXiv:2006.10518}, 2020.

\bibitem{ronneberger2015u}
O.~Ronneberger, P.~Fischer, and T.~Brox, ``U-net: Convolutional networks for biomedical image segmentation,'' in \emph{Medical Image Computing and Computer-Assisted Intervention--MICCAI 2015: 18th International Conference, Munich, Germany, October 5-9, 2015, Proceedings, Part III 18}.\hskip 1em plus 0.5em minus 0.4em\relax Springer, 2015, pp. 234--241.

\bibitem{Peebles2022DiT}
W.~Peebles and S.~Xie, ``Scalable diffusion models with transformers,'' \emph{arXiv preprint arXiv:2212.09748}, 2022.

\bibitem{ma2024deepcache}
X.~Ma, G.~Fang, and X.~Wang, ``Deepcache: Accelerating diffusion models for free,'' in \emph{Proceedings of the IEEE/CVF Conference on Computer Vision and Pattern Recognition}, 2024, pp. 15\,762--15\,772.

\bibitem{chen2024delta}
P.~Chen, M.~Shen, P.~Ye, J.~Cao, C.~Tu, C.-S. Bouganis, Y.~Zhao, and T.~Chen, ``Delta-dit: A training-free acceleration method tailored for diffusion transformers,'' \emph{arXiv preprint arXiv:2406.01125}, 2024.

\bibitem{salimans2022progressive}
T.~Salimans and J.~Ho, ``Progressive distillation for fast sampling of diffusion models,'' \emph{arXiv preprint arXiv:2202.00512}, 2022.

\bibitem{Lu2022KnowledgeDO}
C.~Lu, J.~Zhang, Y.~Chu, Z.~Chen, J.~Zhou, F.~Wu, H.~Chen, and H.~Yang, ``Knowledge distillation of transformer-based language models revisited,'' \emph{ArXiv}, vol. abs/2206.14366, 2022.

\bibitem{kim2023bk}
B.-K. Kim, H.-K. Song, T.~Castells, and S.~Choi, ``Bk-sdm: Architecturally compressed stable diffusion for efficient text-to-image generation,'' in \emph{Workshop on Efficient Systems for Foundation Models@ ICML2023}, 2023.

\bibitem{zhang2024laptop}
D.~Zhang, S.~Li, C.~Chen, Q.~Xie, and H.~Lu, ``Laptop-diff: Layer pruning and normalized distillation for compressing diffusion models,'' \emph{arXiv preprint arXiv:2404.11098}, 2024.

\bibitem{fang2023structural}
G.~Fang, X.~Ma, and X.~Wang, ``Structural pruning for diffusion models,'' in \emph{Advances in Neural Information Processing Systems}, 2023.

\bibitem{liu2024dilatequant}
X.~Liu, Z.~Li, and Q.~Gu, ``Dilatequant: Accurate and efficient diffusion quantization via weight dilation,'' \emph{arXiv preprint arXiv:2409.14307}, 2024.

\bibitem{wang2024quest}
H.~Wang, Y.~Shang, Z.~Yuan, J.~Wu, and Y.~Yan, ``Quest: Low-bit diffusion model quantization via efficient selective finetuning,'' \emph{arXiv preprint arXiv:2402.03666}, 2024.

\bibitem{huang2024tfmq}
Y.~Huang, R.~Gong, J.~Liu, T.~Chen, and X.~Liu, ``Tfmq-dm: Temporal feature maintenance quantization for diffusion models,'' in \emph{Proceedings of the IEEE/CVF Conference on Computer Vision and Pattern Recognition}, 2024, pp. 7362--7371.

\bibitem{he2023ptqd}
Y.~He, L.~Liu, J.~Liu, W.~Wu, H.~Zhou, and B.~Zhuang, ``Ptqd: Accurate post-training quantization for diffusion models,'' \emph{arXiv preprint arXiv:2305.10657}, 2023.

\bibitem{yao2024timestep}
Y.~Yao, F.~Tian, J.~Chen, H.~Lin, G.~Dai, Y.~Liu, and J.~Wang, ``Timestep-aware correction for quantized diffusion models,'' in \emph{European Conference on Computer Vision}.\hskip 1em plus 0.5em minus 0.4em\relax Springer, 2024, pp. 215--232.

\bibitem{huang2024tcaq}
H.~Huang, J.~Chen, J.~Guo, R.~Zhan, and Y.~Wang, ``Tcaq-dm: Timestep-channel adaptive quantization for diffusion models,'' \emph{arXiv preprint arXiv:2412.16700}, 2024.

\bibitem{xiao2023patch}
J.~Xiao, Z.~Li, L.~Yang, and Q.~Gu, ``Patch-wise mixed-precision quantization of vision transformer,'' \emph{arXiv preprint arXiv:2305.06559}, 2023.

\bibitem{li2023vit}
Z.~Li and Q.~Gu, ``I-vit: Integer-only quantization for efficient vision transformer inference,'' in \emph{Proceedings of the IEEE/CVF International Conference on Computer Vision}, 2023, pp. 17\,065--17\,075.

\bibitem{li2023psaq}
Z.~Li, M.~Chen, J.~Xiao, and Q.~Gu, ``Psaq-vit v2: Toward accurate and general data-free quantization for vision transformers,'' \emph{IEEE Transactions on Neural Networks and Learning Systems}, 2023.

\bibitem{chen2021image}
Y.~Chen, L.~Liu, V.~Phonevilay, K.~Gu, R.~Xia, J.~Xie, Q.~Zhang, and K.~Yang, ``Image super-resolution reconstruction based on feature map attention mechanism,'' \emph{Applied Intelligence}, vol.~51, pp. 4367--4380, 2021.

\bibitem{park2018accelerating}
K.~Park and D.-H. Kim, ``Accelerating image classification using feature map similarity in convolutional neural networks,'' \emph{Applied Sciences}, vol.~9, no.~1, p. 108, 2018.

\bibitem{10173725}
D.~Wu, Z.~Guo, A.~Li, C.~Yu, C.~Gao, and N.~Sang, ``Conditional boundary loss for semantic segmentation,'' \emph{IEEE Transactions on Image Processing}, vol.~32, pp. 3717--3731, 2023.

\bibitem{li2023hard}
H.~Li, X.~Wu, F.~Lv, D.~Liao, T.~H. Li, Y.~Zhang, B.~Han, and M.~Tan, ``Hard sample matters a lot in zero-shot quantization,'' in \emph{Proceedings of the IEEE/CVF Conference on Computer Vision and Pattern Recognition}, 2023, pp. 24\,417--24\,426.

\bibitem{botev2017practical}
A.~Botev, H.~Ritter, and D.~Barber, ``Practical gauss-newton optimisation for deep learning,'' in \emph{International Conference on Machine Learning}.\hskip 1em plus 0.5em minus 0.4em\relax PMLR, 2017, pp. 557--565.

\bibitem{yu2015lsun}
F.~Yu, A.~Seff, Y.~Zhang, S.~Song, T.~Funkhouser, and J.~Xiao, ``Lsun: Construction of a large-scale image dataset using deep learning with humans in the loop,'' \emph{arXiv preprint arXiv:1506.03365}, 2015.

\bibitem{heusel2017fid}
M.~Heusel, H.~Ramsauer, T.~Unterthiner, B.~Nessler, and S.~Hochreiter, ``Gans trained by a two time-scale update rule converge to a local nash equilibrium,'' \emph{Advances in neural information processing systems}, vol.~30, 2017.

\bibitem{hessel2021clipscore}
J.~Hessel, A.~Holtzman, M.~Forbes, R.~L. Bras, and Y.~Choi, ``Clipscore: A reference-free evaluation metric for image captioning,'' \emph{arXiv preprint arXiv:2104.08718}, 2021.

\bibitem{salimans2016IS}
T.~Salimans, I.~Goodfellow, W.~Zaremba, V.~Cheung, A.~Radford, and X.~Chen, ``Improved techniques for training gans,'' \emph{Advances in neural information processing systems}, vol.~29, 2016.

\end{thebibliography}
% \vfill
% \vspace{500pt}
% \newpage
 
\vspace{-11pt}

\begin{IEEEbiography}[{\includegraphics[width=1in,height=1.25in,clip,keepaspectratio]{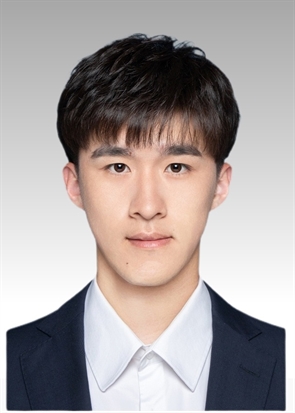}}]{Xuewen Liu} received the B.Sc. degree from China University of Geosciences, Wuhan, China, in 2023. He is currently pursuing the Ph.D. degree with the Institute of Automation, Chinese Academy of Sciences, Beijing, China, and with the School of Artificial Intelligence, University of Chinese Academy of Sciences, Beijing, China. His current research interests include computer vision and efficient deep learning.
\end{IEEEbiography}

\vspace{-11pt}

\begin{IEEEbiography}[{\includegraphics[width=1in,height=1.25in,clip,keepaspectratio]{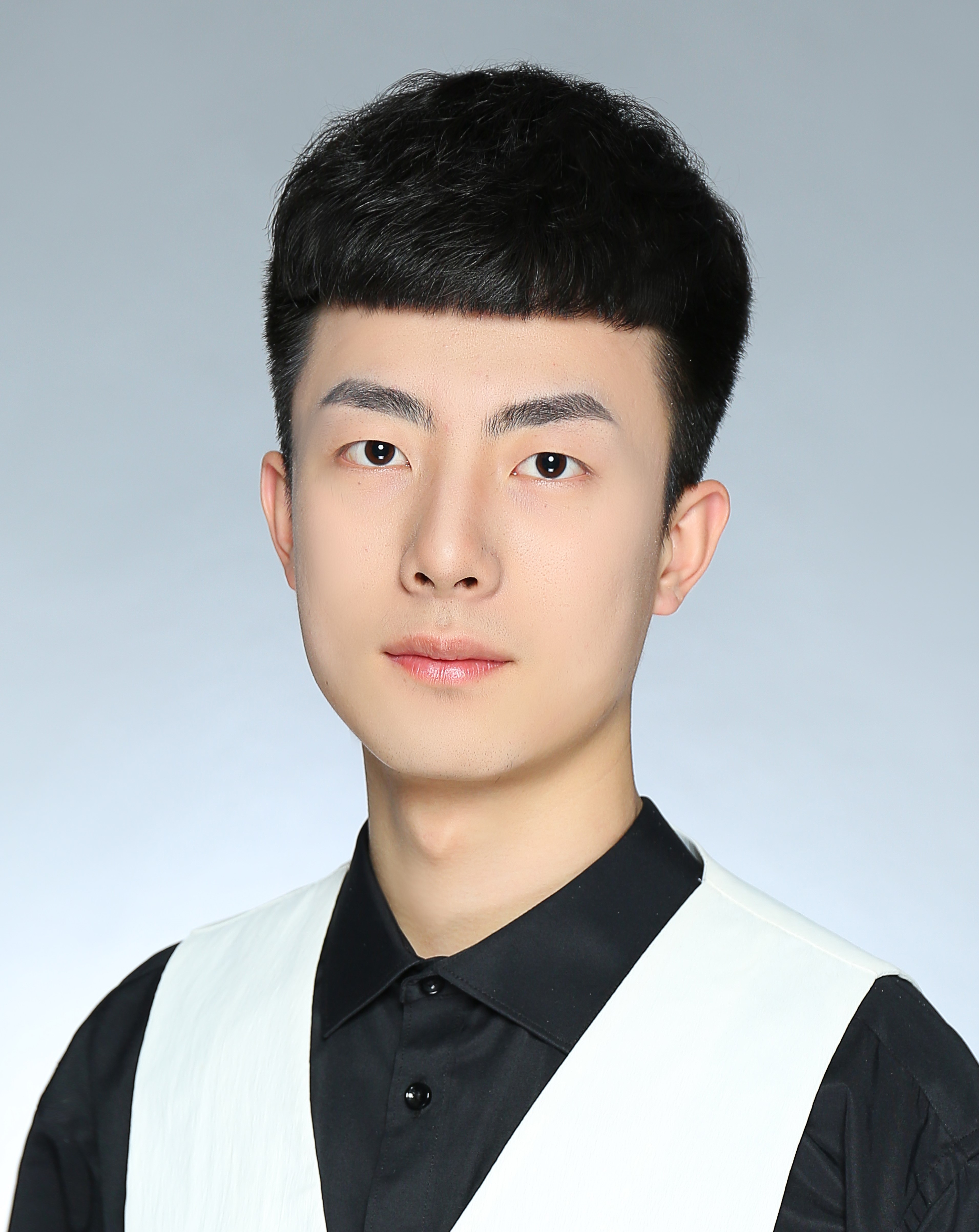}}]{Zhikai Li} received the B.Sc. degree from the Dalian University of Technology, Dalian, China, in 2020. He is currently pursuing the Ph.D. degree with the Institute of Automation, Chinese Academy of Sciences, Beijing, China, and with the School of Artificial Intelligence, University of Chinese Academy of Sciences, Beijing, China. His current research interests include computer vision and
efficient deep learning.
\end{IEEEbiography}

\vspace{-11pt}

\begin{IEEEbiography}[{\includegraphics[width=1in,height=1.25in,clip,keepaspectratio]{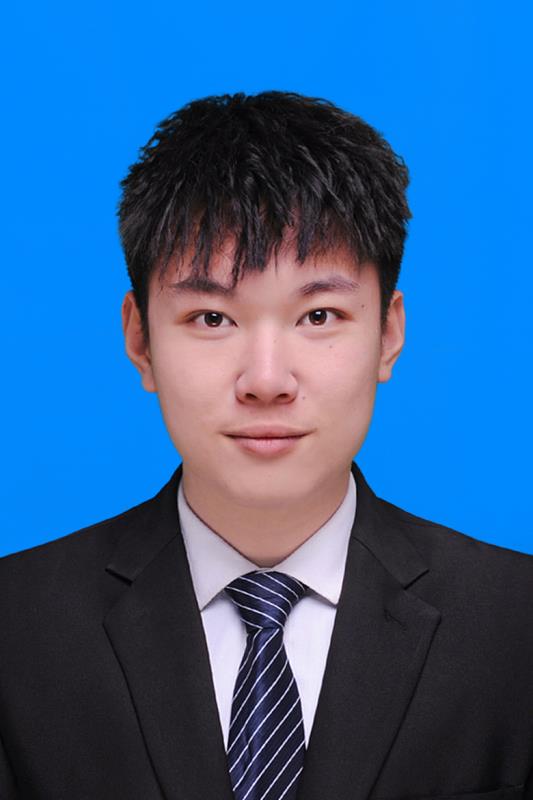}}]{Junrui Xiao} received the B.Sc. degree from Xidian University, Shanxi, China, in 2020. He is currently pursuing the Ph.D. degree with the Institute of Automation, Chinese Academy of Sciences, Beijing, China, and with the School of Artificial Intelligence, University of Chinese Academy of Sciences, Beijing, China. His current research interests include computer vision and model compression.
\end{IEEEbiography}

\vspace{-11pt}

\begin{IEEEbiography}[{\includegraphics[width=1in,height=1.25in,clip,keepaspectratio]{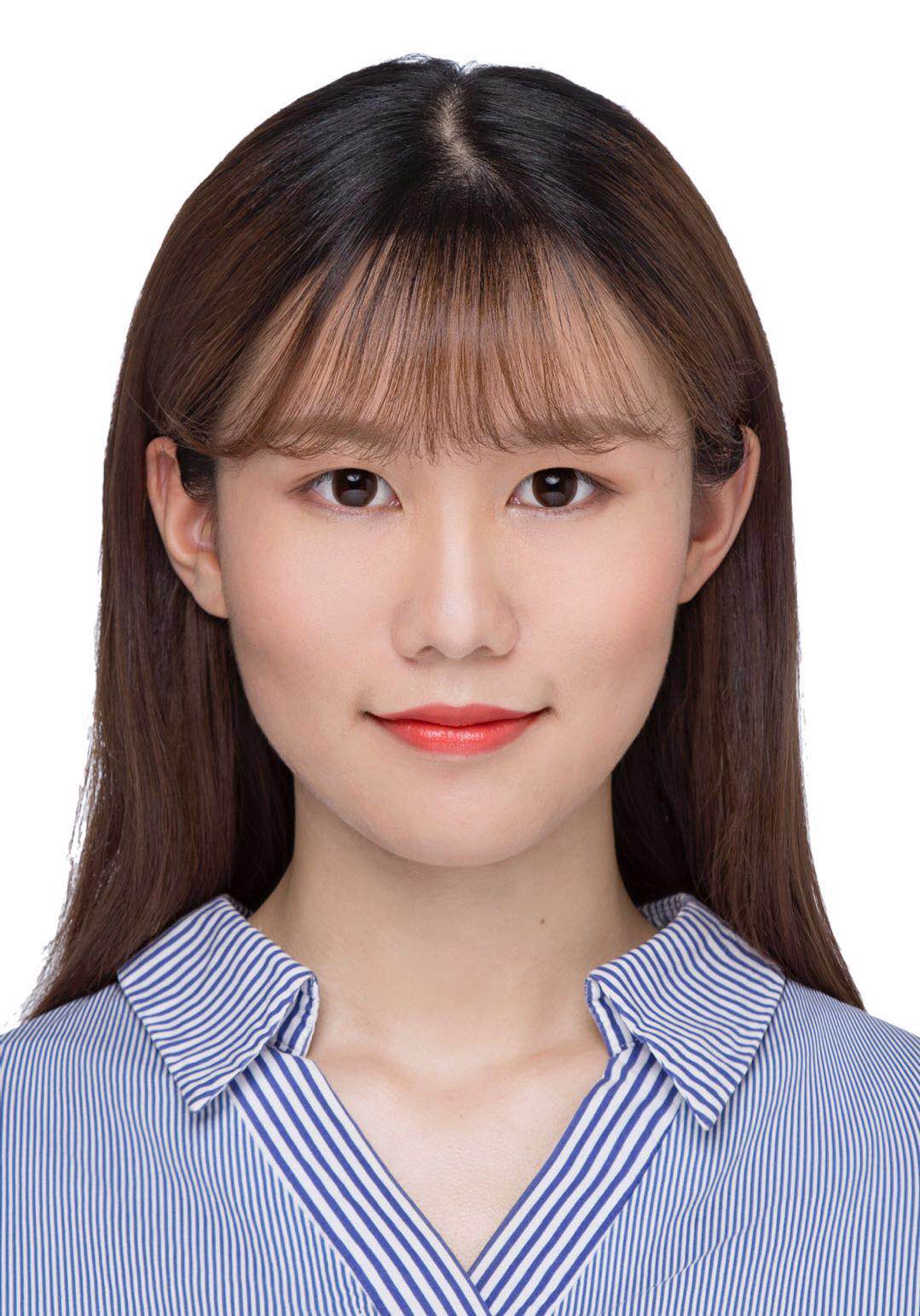}}]{Mengjuan Chen} received the B.E. degree in Automation from the Beijing University of Chemical Technology, Beijing, China, in 2016 and the M.E degree in control engineering from the Institute of Automation, Chinese Academy of Sciences, Beijing, China. She received the Ph.D. degree in Engineering, Hiroshima University, Japan, in 2024. She is currently an assistant professor in Institute of Automation, Chinese Academy of Sciences, China. Her research interests include high-speed image processing and 3D reconstruction.
\end{IEEEbiography}

\vspace{-11pt}

\begin{IEEEbiography}[{\includegraphics[width=1in,height=1.25in,clip,keepaspectratio]{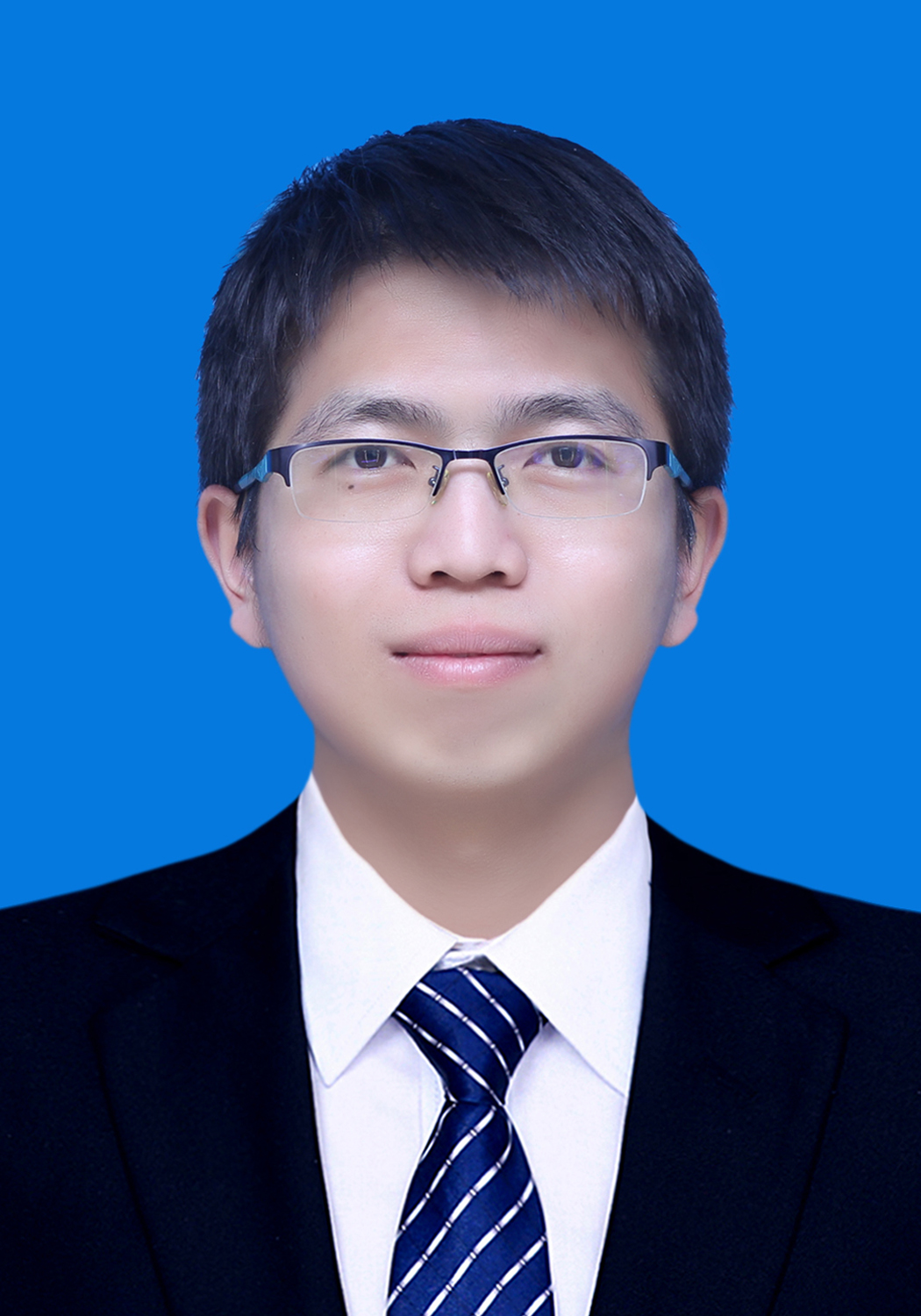}}]{Jianquan Li} received the B.E. degree from Central
South University, Changsha, China, in 2015, and the Ph.D. degree from the Institute of Automation, Chinese Academy of Sciences, Beijing, China, in 2020. He is currently an Assistant Professor with the Institute of Automation, Chinese Academy of Sciences. His primary research interests include industrial visual inspection and algorithm acceleration.
\end{IEEEbiography}

\vspace{-11pt}

\begin{IEEEbiography}[{\includegraphics[width=1in,height=1.25in,clip,keepaspectratio]{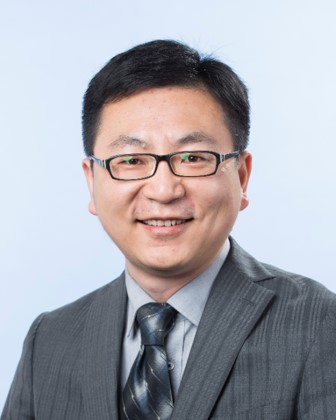}}]{Qingyi GU} (Senior Member, IEEE) received the B.E. degree in Electronic and Information Engineering from Xi’an Jiaotong University, China, in 2005. He received the M.E. degree, and Ph.D. degree in Engineering, Hiroshima University, Japan, in 2010, and 2013 respectively. He is currently a professor in Institute of Automation, Chinese Academy of Sciences, China. His primary research interest is high-speed image processing, and applications in industry and biomedicine.
\end{IEEEbiography}

\vfill

\end{document}